\documentclass[11pt]{article}

\usepackage[final]{acl}

\usepackage{times}
\usepackage{latexsym}
\usepackage[T1]{fontenc}
\usepackage[utf8]{inputenc}
\usepackage{microtype}
\usepackage{inconsolata}
\usepackage{graphicx}
\usepackage{amsmath}
\usepackage{amssymb}
\usepackage{booktabs}
\usepackage{algorithm}
\usepackage{algorithmic}
\usepackage{enumitem}
\usepackage{multirow}
\usepackage{mathtools}
\usepackage{graphicx}
\newcommand{\method}{\textsc{Inspire}}

\title{{\scshape Inspire}: An Internalize-Then-Improve Approach for Example-Driven Mathematical Reasoning}

\author{
  Shuai Wang\textsuperscript{1}\thanks{\ \ indicates equal contribution.},
  Jiayi Kuang\textsuperscript{1*},
  Yinghui Li\textsuperscript{2*},
  Haojing Huang\textsuperscript{2},
  \\
  \textbf{Xinnian Liang\textsuperscript{3}},
  \textbf{Ying Shen\textsuperscript{1,4}\thanks{Corresponding author.}},
  \textbf{Liang Lin\textsuperscript{1}}
  \\[0.4em]
  \textsuperscript{1}Sun Yat-sen University
  \;
  \textsuperscript{2}Tsinghua University
  \;
  \textsuperscript{3}Beihang University
  \;
  \textsuperscript{4}Peng Cheng Laboratory
  \\
  \texttt{wangsh599@mail2.sysu.edu.cn}
  \quad
  \texttt{sheny76@mail.sysu.edu.cn}
}

\begin{document}
\maketitle

\begin{abstract}
Mathematical reasoning has seen rapid progress in large language models (LLMs), yet existing methods optimize predominantly for final-answer correctness, raising the question whether models truly internalize mathematical concepts or merely memorize solution patterns. In human mathematics education, example-based reasoning such as constructing counterexamples to test theorem boundaries reflects deep conceptual understanding, but remains underdeveloped in current LLMs.
Enhancing this capability through preference optimization presents two key challenges: (1) the model's limited example-based reasoning ability makes constructing effective preference pairs inherently difficult; and (2) capability acquisition is progressive, as the model must first learn to adopt this strategy before learning to apply it correctly.
Therefore we propose \method, an Internalize-Then-Improve approach combining Reference-Guided Student Internalization (RGSI), which produces high-quality preference candidates under the policy model's own distribution, with a stage-wise rubric preference training strategy that decomposes learning into method-oriented and correctness-oriented stages. 
Experiments across multiple model scales and families demonstrate consistent improvements, even surpassing larger open-source models, while evaluations on out-of-distribution benchmarks confirm no degradation in general mathematical reasoning ability.
\end{abstract}

\section{Introduction}
Mathematical reasoning is a core aspect of intelligence and a major
focus of large language model (LLM) research. \citep{singh2025gpt5,guo2025deepseekr1,yang2024qwen25math,lu2025youtu}. Enhancing the mathematical reasoning capability of LLMs has become a prominent and fundamental research topic within the community~\citep{xu2026topoagent,zhang2026chatbot,an2026toward,li2026cognitive}.

% Substantial efforts have been devoted to improving the mathematical reasoning ability of LLMs, including scaling up training data through synthesis and augmentation \citep{toshniwal2024openmath,ye2025limo}, supervised fine-tuning on curated mathematical corpora \citep{ying2024internlmmath,muennighoff2025s1}, and reinforcement learning-based post-training that elicits deeper reasoning behaviors \citep{shao2024deepseekmath,du2025kimik15}. Despite their success, most of these methods optimize for final-answer correctness, raising the question of whether current models have truly internalized mathematical concepts or may rely on surface-level solution patterns. 
Substantial recent efforts have been devoted to this goal, including data augmentation \citep{toshniwal2024openmath,ye2025limo}, supervised fine-tuning \citep{ying2024internlmmath,li2025refine,muennighoff2025s1}, and reinforcement learning-based post-training \citep{shao2024deepseekmath,du2025kimik15}. Despite their success, most methods optimize for final-answer correctness, raising the question of whether models truly internalize mathematical concepts or rely on surface-level solution patterns.

\begin{figure}[t]
\centering
\includegraphics[width=\columnwidth]{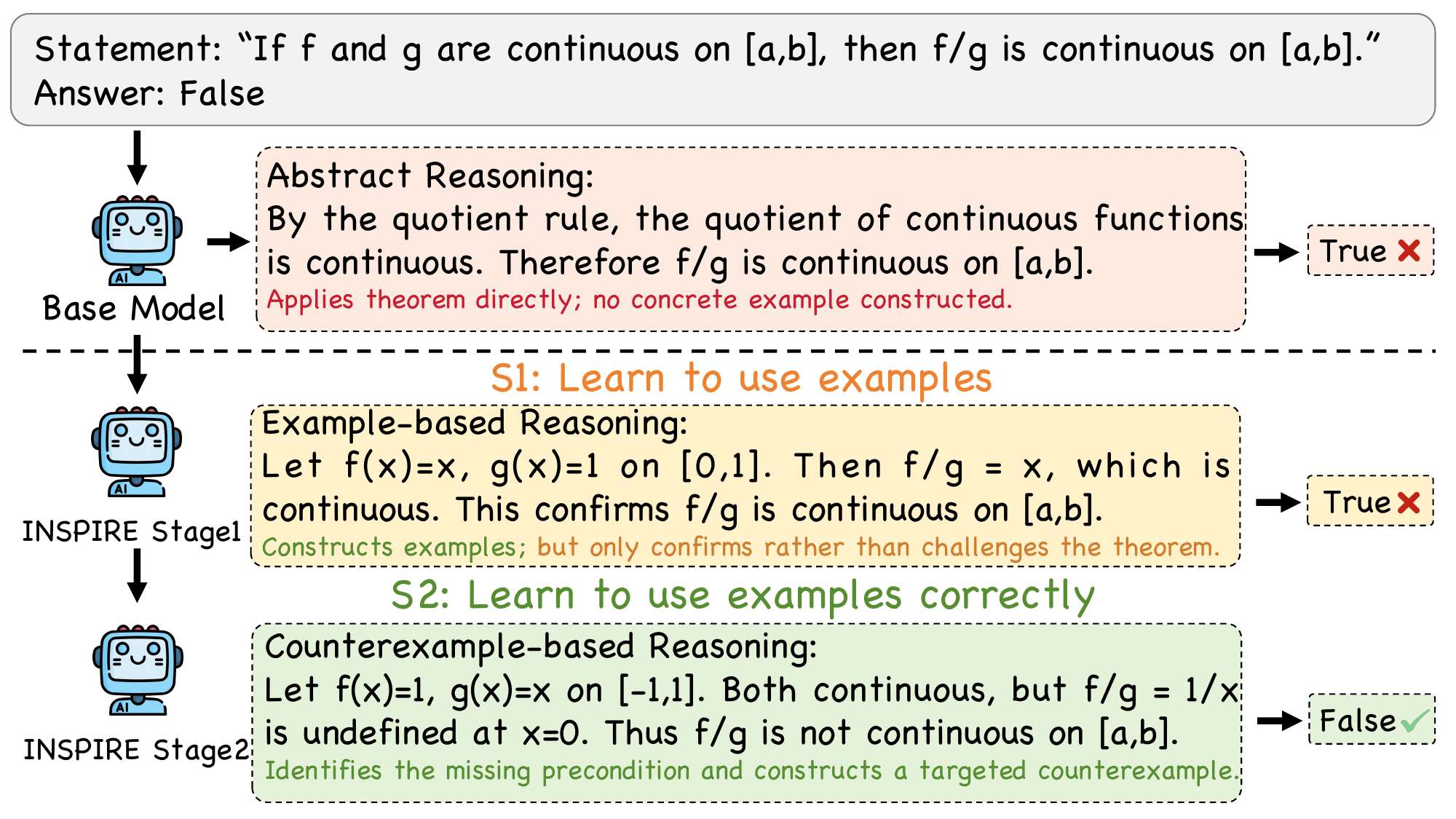}
\caption{Overview of \method's progressive training effect on a mathematical statement requiring example-driven reasoning.}
\label{fig:intro}
\end{figure}

In human mathematics education, example-based reasoning is a fundamental skill reflecting deep conceptual understanding \cite{klymchuk2010counterexamples}. Recent work has shown that this capability remains underdeveloped in current LLMs \cite{wu2024conceptmath, li2025one,kuang2025atomic, li2026learning}. As illustrated in Figure 1, models often rely on abstract theorem application rather than constructing concrete examples to test their reasoning, a tendency reinforced by training paradigms that optimize predominantly for final-answer correctness.

Enhancing this capability through preference optimization presents two key challenges:
\textbf{Data construction under ability scarcity.} 
Since the model exhibits limited example-based reasoning ability, self-sampled candidates are generally low in quality on this dimension, failing to form effective preference pairs. Directly using external references as preferred samples introduces distribution shift in both expression style and reasoning structure, causing the model to replicate surface-level characteristics rather than internalize core analytical strategies.
\textbf{Progressive nature of capability acquisition.} 
Since the model exhibits limited example-based reasoning ability, 
it must first learn to actively adopt this strategy, 
and only then to apply it correctly, making it difficult 
to optimize for both aspects in a single step.

To address these challenges, we propose \method, an \textbf{IN}ternalize-then-improve approach using \textbf{S}tage-wise \textbf{P}reference for \textbf{I}mproved mathematical \textbf{RE}asoning. The internalization stage introduces Reference-Guided Student Internalization (RGSI), which enables the model to comprehend reference reasoning approaches and regenerate responses in its own expression style, producing high-quality preference candidates while avoiding distribution shift. Built upon this, we design a stage-wise preference training strategy that first encourages the model to actively adopt example-based reasoning, and then progressively refines answer correctness on top of the acquired strategy.
Our contributions are summarized as follows:
\begin{itemize}[nosep, leftmargin=*]
    \item We propose \textsc{Inspire}, an Internalize-Then-Improve 
    approach that enhances example-driven mathematical reasoning 
    through preference optimization, addressing both the absence 
    of this ability and its progressive nature of acquisition.
    \item \textsc{Inspire} introduces Reference-Guided Student 
    Internalization to produce high-quality preference 
    candidates, and a stage-wise rubric preference training strategy that decomposes learning into method-oriented 
    and correctness-oriented.
    \item Experiments across multiple model scales and families demonstrate consistent improvements, with comprehensive ablations and analyses validating each design choice.
    Our code has been released at \url{https://github.com/Shea-code-xxx/INSPIRE-code}.
\end{itemize}
% Our contributions are summarized as follows:
% \begin{itemize}
%     \item We propose Reference-Guided Student Internalization (RGSI) along with a multi-dimensional evaluation rubric for constructing high-quality preference pairs under the policy model's own distribution.
%     \item We introduce a stage-wise preference training framework that decomposes the learning process into method-oriented and correctness-oriented stages, enabling progressive acquisition of example-based reasoning capability.
%     \item Experimental results on both 7B and 1.5B model scales demonstrate consistent improvements, and evaluations on out-of-distribution benchmarks confirm no degradation in general mathematical reasoning ability.
% \end{itemize}

\section{Related Work}
\noindent\textbf{Mathematical Reasoning in LLMs.}
Existing approaches generally follow two lines of work.
The first leverages supervised fine-tuning and data augmentation to strengthen problem-solving ability, scaling up training data through instruction synthesis, question rephrasing, and corpus expansion \citep{luo2025wizardmath,yu2024metamath,li2024numinamath}. 
The second employs reinforcement learning for post-training to elicit deeper reasoning capabilities. Representative efforts include GRPO \citep{shao2024deepseekmath}, OpenAI o1 \citep{jaech2024o1}, DeepSeek-R1 \citep{guo2025deepseekr1}, and DAPO \citep{yu2025dapo}, which demonstrate that large-scale reinforcement learning can induce self-verification and reflection behaviors in models.
Despite their success, these methods uniformly optimize for final-answer correctness, paying limited attention to deeper reasoning skills such as concept understanding and discrimination through examples and counterexamples \citep{klymchuk2010counterexamples}. Recent benchmarks such as ConceptMath \citep{wu2024conceptmath} and CounterMath \citep{li2025one} reveal considerable room for improvement in concept-level mathematical reasoning, yet how to effectively enhance this ability through targeted training
that guides models to actively adopt analytical strategies such as example-based reasoning remains largely unexplored.

\noindent\textbf{Preference Optimization for Reasoning.}
Direct Preference Optimization (DPO; \citealp{rafailov2023dpo}) and its variants such as IPO \citep{azar2024general} and SimPO \citep{meng2024simpo} have been applied to mathematical reasoning.
Step-DPO \citep{lai2024stepdpo} and Full-Step-DPO \citep{xu2025fullstepdpo} treat individual reasoning steps as the basic unit for preference optimization, achieving notable results on long-chain reasoning tasks. Math-Shepherd \citep{wang2024mathshepherd} trains a process reward model to provide step-level supervision, moving beyond binary outcome-level signals.
More recently, rubric-based approaches \cite{gunjal2025rubrics, huang2025rubric} have shown that decomposing evaluation into multiple structured criteria can provide richer training signals than single-dimensional rewards. 
A key challenge is constructing high-quality preference data under the policy model's own distribution. Common strategies include rejection sampling  \citep{singh2024beyond,yuan2024selfrewarding} and iterative online DPO that refreshes training pairs with updated model outputs \citep{xu2024chatglmmath}. 
However, when the model has limited target reasoning ability, on-policy sampling alone cannot produce sufficient quality variation, while off-policy data from stronger models risks distribution shift. Moreover, these methods generally rely on answer correctness as the sole preference signal, leaving multi-dimensional evaluation and progressive training strategies largely unexplored.

\begin{figure*}[t]
\centering
\includegraphics[width=0.9\textwidth]{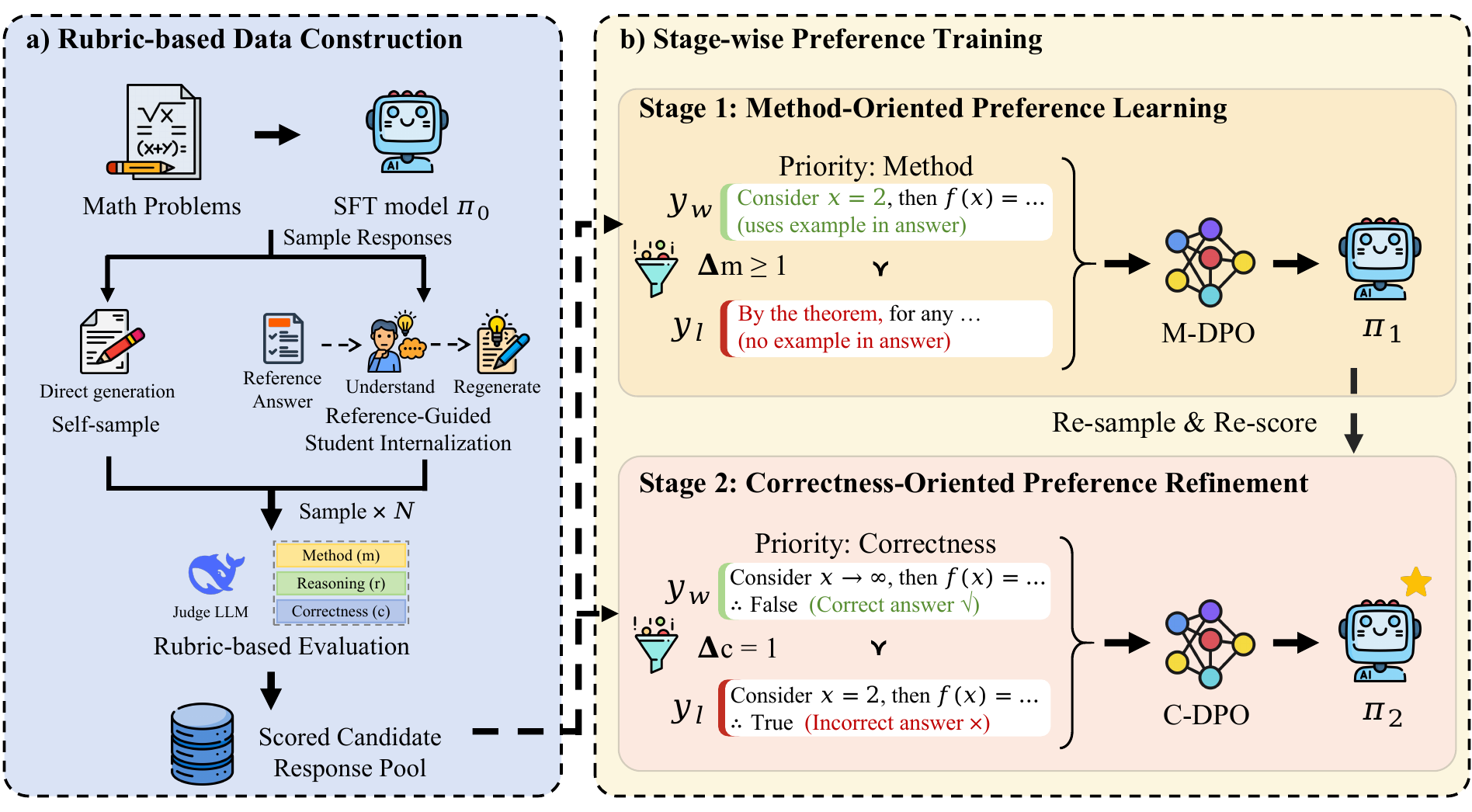}
\caption{Overview of \method. 
(a)~\textbf{Rubric-based Data Construction}: 
candidate responses are obtained via both 
self-sampling and Reference-Guided Student 
Internalization (RGSI) from the base model, 
then scored by a judge LLM along three dimensions 
(Method, Reasoning, Correctness). 
(b)~\textbf{Stage-wise Preference Training}: 
Stage~1 constructs preference pairs prioritizing 
the method dimension to encourage example-based 
reasoning; Stage~2 re-samples from the Stage~1 
model and constructs pairs prioritizing answer 
correctness, with Stage~1 replay samples mixed 
in to prevent forgetting. Note that both 
responses in Stage~2 employ examples, reflecting 
the method acquisition from Stage~1; the 
preference signal now targets correctness.}
\label{fig:pipeline}
\end{figure*}

\section{Methodology}
In this section, we present \textsc{Inspire}
in detail. Built around a multi-dimensional evaluation 
rubric, our method first constructs high-quality preference 
data under the policy model's own distribution 
(\S\ref{sec:data_construction}), and then applies a 
stage-wise training strategy that progressively acquires 
the target reasoning capability (\S\ref{sec:training}). 
Figure~\ref{fig:pipeline} illustrates the overall approach.

\subsection{Rubric-based Preference Data Construction}
\label{sec:data_construction}
Constructing high-quality preference data requires addressing two issues: how to evaluate response quality to distinguish preferred from dispreferred samples, and how to obtain sufficient training pairs with meaningful quality differences. We describe our multi-dimensional evaluation rubrics and data construction strategy in turn.

\begin{table}[t]
\centering
\small
\begin{tabular}{lll}
\toprule
\textbf{Dimension} & \textbf{Range} & \textbf{Description} \\
\midrule
Method ($m$) & 0--2 & Use of example-based analysis \\
Reasoning ($r$) & 0--3 & Logical quality of reasoning \\
Correctness ($c$) & 0--1 & Correctness of the final answer \\
\bottomrule
\end{tabular}
\caption{Three-dimensional evaluation rubric.}
\label{tab:rubric}
\end{table}

\noindent\textbf{Evaluation Rubric.}
Most existing preference optimization methods for 
mathematical reasoning adopt answer correctness as 
the sole preference signal. However, a single 
correctness signal is insufficient for our setting, 
as it cannot distinguish whether the model actively 
adopts example-based reasoning or merely arrives at 
the answer through abstract theorem application.
Inspired by recent rubric-based training approaches 
\citep{gunjal2025rubrics,huang2025rubric} that 
demonstrate the effectiveness of decomposing 
evaluation into multiple structured criteria, and 
drawing on standard practices in mathematics 
education where student proofs are assessed from 
multiple perspectives, we design a three-dimensional 
rubric: \textbf{Method} (whether the response adopts 
an example-based analytical strategy), \textbf{Reasoning} 
(the logical coherence and completeness of the argument), 
and \textbf{Correctness} (whether the final answer 
is correct). We design dimension-specific scoring criteria, 
as summarized in Table~\ref{tab:rubric}
(detailed in Appendix~\ref{app:rubric_detail}). 
Following the LLM-as-a-Judge paradigm 
\citep{zheng2023judging,li2024llms,li2025rethinking,kuang2026process,liu2026tangrampuzzle}, we employ DeepseekV3.2 \citep{liu2024deepseekv3} to score each response independently along 
the three dimensions. The full prompts 
are provided in Appendix~\ref{app:rubric}.

\noindent\textbf{Response Augmentation via Reference-Guided Student Internalization.}
Effective preference learning requires candidate 
responses with meaningful quality differences. 
Meanwhile, preference optimization methods such as 
DPO benefit from training data generated under the 
policy model's own distribution 
\citep{rafailov2023dpo,li2022past,xu2024chatglmmath,shen2026deep}, as 
off-policy data introduces distribution mismatch 
that degrades training effectiveness.
These two requirements create a tension in our setting: 
since the base model exhibits limited example-based reasoning ability, 
on-policy sampling alone produces candidates that 
are uniformly low in quality on this dimension, 
failing to provide sufficient quality gradients 
for preference pair construction.
To resolve this tension, we propose 
\textbf{Reference-Guided Student Internalization 
(RGSI)}. In contrast to standard self-sampling 
where candidates are drawn from $\pi_0(\cdot \mid x)$, 
RGSI conditions the generation on both the problem 
$x$ and the reference solution $r$:
\begin{equation}
  \tilde{y} \sim \pi_0(\cdot \mid x, r),
  \label{eq:rgsi}
\end{equation}
where $r$ provides reasoning guidance while the 
policy model $\pi_0$ serves as the generator, 
ensuring that the output remains on-policy in 
distribution. Meanwhile, informed by the reference, 
the generated responses exhibit higher quality in 
method usage and reasoning than purely self-sampled 
candidates, providing the necessary quality gradients.
Implementation details 
are provided in Appendix~\ref{app:rgsi}. We validate 
the effectiveness of this strategy through ablation 
studies in Section~\ref{sec:ablation}.

\subsection{Stage-wise Rubric Preference Training}
\label{sec:training}
We build our training strategy upon Direct Preference Optimization 
\citep[DPO;][]{rafailov2023dpo}. Since the base model lacks 
example-based reasoning ability, jointly optimizing for method 
acquisition and answer correctness in a single stage is suboptimal 
(verified empirically in Section~\ref{sec:ablation}). We therefore 
decompose training into two stages: \textbf{M-DPO} (Method-oriented 
DPO), which encourages the model to actively adopt example-based 
reasoning, and \textbf{C-DPO} (Correctness-oriented DPO), which 
refines answer correctness based on the acquired example-based 
reasoning strategy.
 
\paragraph{Stage 1:Method-Oriented Preference Learning}
The first stage aims to encourage the model to 
actively adopt example-based reasoning. We construct 
preference pairs primarily based on the method 
dimension. Specifically, for the set of candidate 
responses to each problem, a preference pair 
$(y_w, y_l)$ must satisfy:
\begin{equation}
\label{eq:stage1_pair}
\begin{cases}
m(y_w) - m(y_l) \geq 1 \\
r(y_w) \geq 1,\; r(y_l) \geq 1
\end{cases},
\end{equation}
where $m(\cdot)$ and $r(\cdot)$ denote the method 
and reasoning scores, respectively. The first 
constraint ensures a clear separation between 
responses that employ example-based reasoning and 
those that do not. The latter two serve as quality 
gates, filtering out responses with incoherent 
reasoning. Among all valid pairs, we further 
prioritize those with higher $r(y_w)$, exposing the 
model to high-quality reasoning exemplars while 
acquiring the target strategy. The M-DPO training 
objective is:
\begin{equation}
\label{eq:mdpo}
\scalebox{0.80}{$\displaystyle
\begin{gathered}
\mathcal{L}_{\text{M-DPO}} = -\mathbb{E}_{(x, y_w, y_l) \sim \mathcal{D}_1} \bigg[ \log \sigma \bigg(
\beta_m \log \frac{\pi_\theta(y_w \mid x)}{\pi_0(y_w \mid x)} \\[4pt]
\hspace{10em} -\, \beta_m \log \frac{\pi_\theta(y_l \mid x)}{\pi_0(y_l \mid x)} \bigg) \bigg]\,,
\end{gathered}
$}
\end{equation}
where $\pi_0$ is the SFT model serving as both the 
initialization and the reference policy, $\beta_m$ is 
the temperature parameter, and $\mathcal{D}_1$ denotes 
the set of method-oriented preference pairs. The 
resulting model $\pi_1$ is used as the starting point 
for Stage~2.

\paragraph{Stage 2:Correctness-Oriented Preference Refinement.}
After Stage~1, the model has acquired a preliminary 
ability to employ example-based reasoning. The second 
stage shifts the optimization focus toward answer 
correctness. We first use $\pi_1$ to re-sample 
candidate responses for each problem and score them 
using the same evaluation rubric.
 
The core training signal comes from correctness pairs, 
which establish a direct preference between correct 
and incorrect responses:
\begin{equation}
\label{eq:stage2_pair}
\begin{cases}
c(y_w) = 1, \quad c(y_l) = 0 \\
r(y_w) \geq 1
\end{cases},
\end{equation}
where $c(\cdot)$ denotes the correctness score. The 
reasoning constraint $r(y_w) \geq 1$ filters out 
responses that arrive at the correct answer through 
flawed reasoning. As a complementary signal, when 
multiple candidates are all correct, we further 
construct quality pairs satisfying $c(y_w) = c(y_l) 
= 1$ and $r(y_w) - r(y_l) \geq 1$, encouraging the 
model to produce more rigorous arguments among 
correct solutions. For both types, the method score 
serves as a secondary ranking criterion, continuously 
reinforcing the strategy acquired in Stage~1. 
The C-DPO training objective is:
\begin{equation}
\label{eq:cdpo}
\scalebox{0.80}{$\displaystyle
\begin{gathered}
\mathcal{L}_{\text{C-DPO}} = -\mathbb{E}_{(x, y_w, y_l) \sim \mathcal{D}_2} \bigg[ \log \sigma \bigg(
\beta_c \log \frac{\pi_\theta(y_w \mid x)}{\pi_1(y_w \mid x)} \\[4pt]
\hspace{10em} -\, \beta_c \log \frac{\pi_\theta(y_l \mid x)}{\pi_1(y_l \mid x)} \bigg) \bigg]\,,
\end{gathered}
$}
\end{equation}
where $\pi_1$ serves as both the initialization and 
the reference policy, $\beta_c$ is the temperature 
parameter, and $\mathcal{D}_2$ denotes the 
correctness-oriented preference pairs augmented with 
a subset of Stage~1 pairs to prevent forgetting of 
the previously acquired reasoning strategy.
The complete training pipeline is summarized in 
Algorithm~\ref{alg:pipeline}.

\begin{algorithm}[!b]
\caption{Stage-wise Rubric Preference Training}
\label{alg:pipeline}
\begin{algorithmic}[1]
\REQUIRE Problem set $\mathcal{D}$, base model $\pi_0$, ground-truth solutions $\mathcal{G}$, judge model $\mathcal{J}$
\ENSURE Final model $\pi_2$
\STATE \textcolor{gray}{\textit{// Data Preparation}}
\FOR{each problem $d \in \mathcal{D}$}
    \STATE Sample $K$ responses from $\pi_0$ given $d$
    \STATE Generate rewrite responses from $\pi_0$ guided by $\mathcal{G}(d)$
    \STATE Score all responses with $\mathcal{J}$ on $(m, r, c)$
\ENDFOR
\STATE \textcolor{gray}{\textit{// Stage 1: M-DPO (Method-Oriented Preference Learning)}}
\FOR{each problem $d \in \mathcal{D}$}
    \STATE Construct pairs $\mathcal{D}_1(d)$ where $m(y_w) - m(y_l) \geq 1$, 
    $r(y_w) \geq 1$, $r(y_l) \geq 1$; prioritize pairs with higher $r(y_w)$
\ENDFOR
\STATE Train $\pi_1 \leftarrow \text{M-DPO}(\pi_0,\; \bigcup_x \mathcal{D}_1(d))$
\STATE \textcolor{gray}{\textit{// Stage 2: C-DPO (Correctness-Oriented Preference Refinement)}}
\FOR{each problem $d \in \mathcal{D}$}
    \STATE Re-sample $K$ responses from $\pi_1$ given $d$
    \STATE Score all responses with $\mathcal{J}$ on $(m, r, c)$
    \STATE Construct correctness pairs: $c(y_w){=}1$ vs $c(y_l){=}0$
    \STATE Construct quality pairs: both $c{=}1$, $r(y_w) - r(y_l) \geq 1$
\ENDFOR
\STATE $\mathcal{D}_2 \leftarrow \bigcup_x \mathcal{D}_2(x) \;\cup\; \text{Replay}(\mathcal{D}_1)$
\STATE Train $\pi_2 \leftarrow \text{C-DPO}(\pi_1,\; \mathcal{D}_2)$
\RETURN $\pi_2$
\end{algorithmic}
\end{algorithm}

\section{Experiments}
\subsection{Experimental Setup}
\paragraph{Dataset.}
We select mathematical proof problems involving examples, 
counterexamples, or constructions from the training set of 
BrokenMath~\citep{petrov2025brokenmath}, a theorem-proving 
benchmark of approximately 15K problems sourced from 
DeepTheorem and NuminaMath-1.5, as our training data.
Candidate problems are identified via DeepSeek-R1 based on 
predefined criteria (Appendix~\ref{app:filtering})
and manually verified, yielding 1,275 problems.
For each problem, we sample 16 candidate responses from the 
base model and generate 4 additional responses via RGSI. 
After preference pair construction, Stage~1 and Stage~2 each 
yield approximately 14{,}000 training pairs.
% The SFT stage is trained on the ground-truth reference of the problems, serving as a cold start before preference training.
The SFT stage is trained on ground-truth references as a cold start.
We evaluate on CounterMath~\citep{li2025one}, 
which directly targets
example-driven reasoning, using its four metrics: Macro-F1, Examples,
Strict Align, and Loose Align (Appendix~\ref{app:eval-details}).

\noindent\textbf{Baselines.}
We compare against a broad range of models, including commercial models (e.g., Claude-4.5-Sonnet \citep{anthropic2025claude45sonnet}, Gemini~3 Pro \citep{google2025gemini3pro}) and open-source models spanning 3B to 72B parameters, covering both general-purpose and math-specialized models. 
% All baselines and their references are listed in Table~\ref{tab:main}. To demonstrate the contribution of each component, we report progressive results on all three model configurations: Base, +SFT, +Stage~1, and +Stage~1+Stage~2. 
All baselines are listed in Table~\ref{tab:main}. We report progressive results (Base, +SFT, +Stage 1, +Stage 1+Stage 2) on all three configurations.
This setup covers two model scales within the Qwen2.5-Math (1.5/7B) \citep{yang2024qwen25math}  and Llama-3.1-8B-Instruct \citep{grattafiori2024llama3} as a cross-family configuration, allowing us to assess both scale and cross-family generalizability.

\paragraph{Implementation Details.}
We use Qwen2.5-Math-7B-Instruct as our 
primary base model and conduct all training with 
LLaMA-Factory~\citep{zheng2024llamafactory}. 
Both the SFT and DPO stages employ LoRA~\citep{hu2022lora}. 
DeepSeek-V3.2 serves as the judge model for rubric scoring.
During inference, we use vLLM~\citep{kwon2023vllm} for generation with greedy decoding for all evaluations. 
All experiments are conducted on 4 NVIDIA L20 GPUs. 
We additionally apply the same pipeline to Qwen2.5-Math-1.5B-Instruct with identical hyperparameters. 
To further verify cross-family generalizability, we apply the same pipeline and hyperparameters to Llama-3.1-8B-Instruct.
Full hyperparameter settings are provided in Appendix~\ref{app:eval-details}.

\begin{table*}[t]
\centering
\small
\caption{Main evaluation results on CounterMath. Results are grouped into commercial models, open-source models, and our method with progressive training stages. Bold indicates the best result within each group.}
\label{tab:main}
\begin{tabular}{l|c@{\hspace{1.4em}}|ccc@{\hspace{1.6em}}}
\toprule
\multirow{2}{*}{\textbf{Models}} & \textbf{Judgement} & \multicolumn{3}{c}{\textbf{Rationale Reasoning}} \\
& \textbf{F1 (macro)} & \textbf{Examples (\%)} & \textbf{Strict (\%)} & \textbf{Loose (\%)} \\
\midrule
\multicolumn{5}{c}{\textit{Commercial models}} \\
\midrule
KIMI-K2.5 \citep{bai2026kimik25} & 51.79 & 71.63 & 34.78 & 45.80 \\
GLM-4.6 \citep{zeng2025glm}& 58.39 & 73.19 & 41.45 & 45.06 \\
Claude-4.5-Sonnet \citep{anthropic2025claude45sonnet} & 74.24 & 72.94 & 39.30 & 45.64 \\
DeepSeek-R1 \citep{guo2025deepseekr1} & 73.50 & 80.59 & 42.52 & 46.95 \\
Gemini-3-Pro \citep{google2025gemini3pro} & \textbf{80.43} & \textbf{81.17} & \textbf{45.56} & \textbf{50.82} \\
\midrule
\multicolumn{5}{c}{\textit{Open-source models}} \\
\midrule
Llama-3.2-3B-Instruct \citep{grattafiori2024llama3} & 35.78 & 66.53 & 4.28 & 4.76 \\
Phi-4-mini-3.8B-instruct \citep{abouelenin2025phi4mini} & 29.67 & 49.51 & 7.81 & 9.04 \\
Qwen3-4B \citep{yang2025qwen3} & 33.95 & 58.14 & 14.47 & 17.93 \\
DeepSeek-Math-7B-rl \citep{shao2024deepseekmath} & 33.91 & 65.46 & 10.19 & 10.94 \\
Eurus-2-7B-PRIME \citep{cui2025prime} & 36.75 & 62.83 & 14.63 & 17.35 \\
Qwen3-32B \citep{yang2025qwen3} & \textbf{45.93} & \textbf{76.97} & \textbf{25.41} & \textbf{30.67} \\
Qwen2.5-Math-72B-Instruct \citep{yang2024qwen25math} & 41.89 & 75.99 & 21.46 & 24.67 \\
\midrule
\multicolumn{5}{c}{\textit{INSPIRE (Ours)}} \\
% \midrule
% Llama-3.1-8B-Instruct \citep{grattafiori2024llama3} & 41.28 & 76.73 & 6.57 & 7.56 \\
% \quad + SFT & 42.98 & 80.10 & 7.40 & 8.31 \\
% \quad + Stage 1 & 48.15 & 89.88 & 8.22 & 9.78 \\
% \quad + Stage 1 + Stage 2 (Full) & \textbf{50.27}\rlap{\,\scalebox{0.65}{\textcolor{darkgray}{(+8.99)}}} & \textbf{93.09}\rlap{\,\scalebox{0.65}{\textcolor{darkgray}{(+16.36)}}} & \textbf{9.45}\rlap{\,\scalebox{0.65}{\textcolor{darkgray}{(+2.88)}}} & \textbf{11.80}\rlap{\,\scalebox{0.65}{\textcolor{darkgray}{(+4.24)}}} \\
% \midrule
% Qwen2.5-Math-1.5B-Instruct \citep{yang2024qwen25math} & 35.71 & 69.24 & 9.04 & 10.60 \\
% \quad + SFT & 36.36 & 69.82 & 9.45 & 10.93 \\
% \quad + Stage 1 & 37.26 & 74.84 & 10.85 & 11.84 \\
% \quad + Stage 1 + Stage 2 (Full) & \textbf{39.02}\rlap{\,\scalebox{0.65}{\textcolor{darkgray}{(+3.31)}}} & \textbf{76.32}\rlap{\,\scalebox{0.65}{\textcolor{darkgray}{(+7.08)}}} & \textbf{11.68}\rlap{\,\scalebox{0.65}{\textcolor{darkgray}{(+2.64)}}} & \textbf{13.49}\rlap{\,\scalebox{0.65}{\textcolor{darkgray}{(+2.89)}}} \\
% \midrule
% Qwen2.5-Math-7B-Instruct \citep{yang2024qwen25math} & 38.67 & 75.25 & 14.31 & 16.61 \\
% \quad + SFT & 39.06 & 77.63 & 14.39 & 17.02 \\
% \quad + Stage 1 & 43.05 & 81.33 & 16.53 & 19.08 \\
% \quad + Stage 1 + Stage 2 (Full) & \textbf{45.91}\rlap{\,\scalebox{0.65}{\textcolor{darkgray}{(+7.24)}}} & \textbf{84.79}\rlap{\,\scalebox{0.65}{\textcolor{darkgray}{(+9.54)}}} & \textbf{17.30}\rlap{\,\scalebox{0.65}{\textcolor{darkgray}{(+2.99)}}} & \textbf{20.07}\rlap{\,\scalebox{0.65}{\textcolor{darkgray}{(+3.46)}}} \\
% \bottomrule
\midrule
Llama-3.1-8B-Instruct \citep{grattafiori2024llama3} & 41.28 & 76.73 & 6.57 & 7.56 \\
\quad + SFT & 42.98\rlap{\,\scalebox{0.65}{\textcolor{darkgray}{(+1.70)}}} & 80.10\rlap{\,\scalebox{0.65}{\textcolor{darkgray}{(+3.37)}}} & 7.40\rlap{\,\scalebox{0.65}{\textcolor{darkgray}{(+0.83)}}} & 8.31\rlap{\,\scalebox{0.65}{\textcolor{darkgray}{(+0.75)}}} \\
\quad + Stage 1 & 48.15\rlap{\,\scalebox{0.65}{\textcolor{darkgray}{(+6.87)}}} & 89.88\rlap{\,\scalebox{0.65}{\textcolor{darkgray}{(+13.15)}}} & 8.22\rlap{\,\scalebox{0.65}{\textcolor{darkgray}{(+1.65)}}} & 9.78\rlap{\,\scalebox{0.65}{\textcolor{darkgray}{(+2.22)}}} \\
\quad + Stage 1 + Stage 2 (Full) & \textbf{50.27}\rlap{\,\scalebox{0.65}{\textcolor{darkgray}{(+8.99)}}} & \textbf{93.09}\rlap{\,\scalebox{0.65}{\textcolor{darkgray}{(+16.36)}}} & \textbf{9.45}\rlap{\,\scalebox{0.65}{\textcolor{darkgray}{(+2.88)}}} & \textbf{11.80}\rlap{\,\scalebox{0.65}{\textcolor{darkgray}{(+4.24)}}} \\
\midrule
Qwen2.5-Math-1.5B-Instruct \citep{yang2024qwen25math} & 35.71 & 69.24 & 9.04 & 10.60 \\
\quad + SFT & 36.36\rlap{\,\scalebox{0.65}{\textcolor{darkgray}{(+0.65)}}} & 69.82\rlap{\,\scalebox{0.65}{\textcolor{darkgray}{(+0.58)}}} & 9.45\rlap{\,\scalebox{0.65}{\textcolor{darkgray}{(+0.41)}}} & 10.93\rlap{\,\scalebox{0.65}{\textcolor{darkgray}{(+0.33)}}} \\
\quad + Stage 1 & 37.26\rlap{\,\scalebox{0.65}{\textcolor{darkgray}{(+1.55)}}} & 74.84\rlap{\,\scalebox{0.65}{\textcolor{darkgray}{(+5.60)}}} & 10.85\rlap{\,\scalebox{0.65}{\textcolor{darkgray}{(+1.81)}}} & 11.84\rlap{\,\scalebox{0.65}{\textcolor{darkgray}{(+1.24)}}} \\
\quad + Stage 1 + Stage 2 (Full) & \textbf{39.02}\rlap{\,\scalebox{0.65}{\textcolor{darkgray}{(+3.31)}}} & \textbf{76.32}\rlap{\,\scalebox{0.65}{\textcolor{darkgray}{(+7.08)}}} & \textbf{11.68}\rlap{\,\scalebox{0.65}{\textcolor{darkgray}{(+2.64)}}} & \textbf{13.49}\rlap{\,\scalebox{0.65}{\textcolor{darkgray}{(+2.89)}}} \\
\midrule
Qwen2.5-Math-7B-Instruct \citep{yang2024qwen25math} & 38.67 & 75.25 & 14.31 & 16.61 \\
\quad + SFT & 39.06\rlap{\,\scalebox{0.65}{\textcolor{darkgray}{(+0.39)}}} & 77.63\rlap{\,\scalebox{0.65}{\textcolor{darkgray}{(+2.38)}}} & 14.39\rlap{\,\scalebox{0.65}{\textcolor{darkgray}{(+0.08)}}} & 17.02\rlap{\,\scalebox{0.65}{\textcolor{darkgray}{(+0.41)}}} \\
\quad + Stage 1 & 43.05\rlap{\,\scalebox{0.65}{\textcolor{darkgray}{(+4.38)}}} & 81.33\rlap{\,\scalebox{0.65}{\textcolor{darkgray}{(+6.08)}}} & 16.53\rlap{\,\scalebox{0.65}{\textcolor{darkgray}{(+2.22)}}} & 19.08\rlap{\,\scalebox{0.65}{\textcolor{darkgray}{(+2.47)}}} \\
\quad + Stage 1 + Stage 2 (Full) & \textbf{45.91}\rlap{\,\scalebox{0.65}{\textcolor{darkgray}{(+7.24)}}} & \textbf{84.79}\rlap{\,\scalebox{0.65}{\textcolor{darkgray}{(+9.54)}}} & \textbf{17.30}\rlap{\,\scalebox{0.65}{\textcolor{darkgray}{(+2.99)}}} & \textbf{20.07}\rlap{\,\scalebox{0.65}{\textcolor{darkgray}{(+3.46)}}} \\
\bottomrule
\end{tabular}
\end{table*}

\subsection{Main Results}

Table~\ref{tab:main} presents the main results on 
CounterMath. Across all three model configurations, each training stage 
yields consistent improvements. On the 7B scale, Stage~1 
increases Examples from 77.63\% to 81.33\% and F1 from 39.06 
to 43.05, confirming that method-oriented preference learning 
encourages example-based reasoning adoption. Stage~2 further 
raises F1 to 45.91 and Examples to 84.79\%, demonstrating 
effective correctness refinement atop the acquired strategy. 
The alignment metrics improve steadily across stages, though 
absolute scores remain moderate since models may construct 
valid yet distinct examples that differ from references, as 
also noted in the CounterMath analysis, we therefore focus on relative improvement across stages.
The same progressive pattern holds on both the 1.5B scale (F1: +2.66, Ex.: +7.08) and Llama-3.1-8B-Instruct (F1: +7.29, Ex.: +12.99), confirming the consistency of our approach across model scales and families. SFT provides marginal gains, primarily serving as format-level cold start.

Compared with open-source models, our 7B model \textbf{surpasses 
Qwen2.5-Math-72B-Instruct} (F1: 41.89) and approaches 
Qwen3-32B (F1: 45.93). Notably, smaller models generally 
exhibit lower Examples rates with correspondingly 
lower F1, suggesting that example usage is a key bottleneck 
for this task. More broadly, F1 and Examples are strongly 
correlated across models, confirming that actively employing 
examples is a primary driver of judgment accuracy on 
CounterMath. 
Our models also achieve \textbf{higher Examples rates than all other evaluated models}: 84.79\% (Qwen-7B) and 93.09\% (Llama-8B), both exceeding DeepSeek-R1 (80.59\%).
Furthermore, the 1.5B model after 
full training (F1: 39.02) \textbf{reaches comparable performance to 
the 7B base model} (F1: 38.67), demonstrating that 
our method can effectively compensate for the 
capacity gap through targeted training. The remaining 
gap to commercial models is largely attributable to 
base model capacity.
Ablation studies and analyses in Sections~\ref{sec:ablation} and \ref{sec:analysis} further validate each design choice.

\begin{table*}[t]
\centering
\small
\caption{Out-of-distribution evaluation results on general mathematical reasoning benchmarks. 
Values in parentheses indicate the difference relative to the base model.}
\label{tab:ood}
\begin{tabular}{l|ccccc}
\toprule
\textbf{Models} & \textbf{GSM8K} & \textbf{MATH500} & \textbf{AIME 2024} & \textbf{GAOKAO-mQA} & \textbf{MMLU-cMath} \\
% \midrule
% Llama-3.1-8B-Instruct & 82.94 & 47.80 & 6.67 & 32.48 & 42.00 \\
% + Stage 1 & 83.40 & 49.00 & 6.67 & 34.76 & 43.00 \\
% + Stage 1 + Stage 2 & \textbf{83.47}\rlap{\,\scalebox{0.65}{\textcolor{darkgray}{(+0.53)}}} & \textbf{49.60}\rlap{\,\scalebox{0.65}{\textcolor{darkgray}{(+1.80)}}} & 6.67\rlap{\,\scalebox{0.65}{\textcolor{darkgray}{(+0.00)}}} & \textbf{35.04}\rlap{\,\scalebox{0.65}{\textcolor{darkgray}{(+2.56)}}} & \textbf{45.00}\rlap{\,\scalebox{0.65}{\textcolor{darkgray}{(+3.00)}}} \\
% \midrule
% Qwen2.5-Math-1.5B-Instruct & 85.22 & 73.80 & 10.00 & 73.22 & 70.00 \\
% + Stage 1 & 85.22 & \textbf{75.20} & 10.00 & 73.79 & 70.00 \\
% + Stage 1 + Stage 2 & \textbf{85.75}\rlap{\,\scalebox{0.65}{\textcolor{darkgray}{(+0.53)}}} & 75.00\rlap{\,\scalebox{0.65}{\textcolor{darkgray}{(+1.20)}}} & 10.00\rlap{\,\scalebox{0.65}{\textcolor{darkgray}{(+0.00)}}} & \textbf{74.83}\rlap{\,\scalebox{0.65}{\textcolor{darkgray}{(+1.61)}}} & \textbf{73.00}\rlap{\,\scalebox{0.65}{\textcolor{darkgray}{(+3.00)}}} \\
% \midrule
% Qwen2.5-Math-7B-Instruct & 94.54 & 84.20 & 13.33 & 78.06 & 72.00 \\
% + Stage 1 & 95.45 & 84.80 & 16.67 & 79.77 & 73.00 \\
% + Stage 1 + Stage 2 & \textbf{95.68}\rlap{\,\scalebox{0.65}{\textcolor{darkgray}{(+1.14)}}} & \textbf{84.80}\rlap{\,\scalebox{0.65}{\textcolor{darkgray}{(+0.60)}}} & \textbf{20.00}\rlap{\,\scalebox{0.65}{\textcolor{darkgray}{(+6.67)}}} & \textbf{80.06}\rlap{\,\scalebox{0.65}{\textcolor{darkgray}{(+2.00)}}} & \textbf{75.00}\rlap{\,\scalebox{0.65}{\textcolor{darkgray}{(+3.00)}}} \\
% \bottomrule
\midrule
Llama-3.1-8B-Instruct & 82.94 & 47.80 & 6.67 & 32.48 & 42.00 \\
+ Stage 1 & 83.40\rlap{\,\scalebox{0.65}{\textcolor{darkgray}{(+0.46)}}} & 49.00\rlap{\,\scalebox{0.65}{\textcolor{darkgray}{(+1.20)}}} & 6.67\rlap{\,\scalebox{0.65}{\textcolor{darkgray}{(+0.00)}}} & 34.76\rlap{\,\scalebox{0.65}{\textcolor{darkgray}{(+2.28)}}} & 43.00\rlap{\,\scalebox{0.65}{\textcolor{darkgray}{(+1.00)}}} \\
+ Stage 1 + Stage 2 & \textbf{83.47}\rlap{\,\scalebox{0.65}{\textcolor{darkgray}{(+0.53)}}} & \textbf{49.60}\rlap{\,\scalebox{0.65}{\textcolor{darkgray}{(+1.80)}}} & 6.67\rlap{\,\scalebox{0.65}{\textcolor{darkgray}{(+0.00)}}} & \textbf{35.04}\rlap{\,\scalebox{0.65}{\textcolor{darkgray}{(+2.56)}}} & \textbf{45.00}\rlap{\,\scalebox{0.65}{\textcolor{darkgray}{(+3.00)}}} \\
\midrule
Qwen2.5-Math-1.5B-Instruct & 85.22 & 73.80 & 10.00 & 73.22 & 70.00 \\
+ Stage 1 & 85.22\rlap{\,\scalebox{0.65}{\textcolor{darkgray}{(+0.00)}}} & \textbf{75.20}\rlap{\,\scalebox{0.65}{\textcolor{darkgray}{(+1.40)}}} & 10.00\rlap{\,\scalebox{0.65}{\textcolor{darkgray}{(+0.00)}}} & 73.79\rlap{\,\scalebox{0.65}{\textcolor{darkgray}{(+0.57)}}} & 70.00\rlap{\,\scalebox{0.65}{\textcolor{darkgray}{(+0.00)}}} \\
+ Stage 1 + Stage 2 & \textbf{85.75}\rlap{\,\scalebox{0.65}{\textcolor{darkgray}{(+0.53)}}} & 75.00\rlap{\,\scalebox{0.65}{\textcolor{darkgray}{(+1.20)}}} & 10.00\rlap{\,\scalebox{0.65}{\textcolor{darkgray}{(+0.00)}}} & \textbf{74.83}\rlap{\,\scalebox{0.65}{\textcolor{darkgray}{(+1.61)}}} & \textbf{73.00}\rlap{\,\scalebox{0.65}{\textcolor{darkgray}{(+3.00)}}} \\
\midrule
Qwen2.5-Math-7B-Instruct & 94.54 & 84.20 & 13.33 & 78.06 & 72.00 \\
+ Stage 1 & 95.45\rlap{\,\scalebox{0.65}{\textcolor{darkgray}{(+0.91)}}} & 84.80\rlap{\,\scalebox{0.65}{\textcolor{darkgray}{(+0.60)}}} & 16.67\rlap{\,\scalebox{0.65}{\textcolor{darkgray}{(+3.34)}}} & 79.77\rlap{\,\scalebox{0.65}{\textcolor{darkgray}{(+1.71)}}} & 73.00\rlap{\,\scalebox{0.65}{\textcolor{darkgray}{(+1.00)}}} \\
+ Stage 1 + Stage 2 & \textbf{95.68}\rlap{\,\scalebox{0.65}{\textcolor{darkgray}{(+1.14)}}} & \textbf{84.80}\rlap{\,\scalebox{0.65}{\textcolor{darkgray}{(+0.60)}}} & \textbf{20.00}\rlap{\,\scalebox{0.65}{\textcolor{darkgray}{(+6.67)}}} & \textbf{80.06}\rlap{\,\scalebox{0.65}{\textcolor{darkgray}{(+2.00)}}} & \textbf{75.00}\rlap{\,\scalebox{0.65}{\textcolor{darkgray}{(+3.00)}}} \\
\bottomrule
\end{tabular}
\end{table*}
\subsection{Out-of-Distribution Evaluation}
To examine the broader impact of our training on general mathematical reasoning, we evaluate on several out-of-distribution benchmarks including GSM8K, MATH500, AIME 2024, GAOKAO-mathQA, and MMLU-collegeMath.
% As shown in Table~\ref{tab:ood}, the final model consistently maintains or improves performance across all benchmarks on all three model configurations, suggesting that our specialized training may also benefit general mathematical reasoning through improved analytical strategies.
As shown in Table~\ref{tab:ood}, the final model consistently maintains or improves performance across all benchmarks on all three model configurations. The unchanged AIME 2024 scores for the two smaller configurations are expected given its 30-problem granularity and competition-level difficulty; notably, Qwen2.5-Math-7B achieves +6.67 on this benchmark. These results suggest that our specialized training may also benefit general mathematical reasoning through improved analytical strategies.

% \begin{table}[t]
% \centering
% \small
% \caption{Ablation on training strategy. The first group compares alternative training strategies; the second group ablates individual stages of our pipeline.}
% \label{tab:ablation-stage}
% \begin{tabular}{l|cccc}
% \toprule
% \textbf{Strategy} & \textbf{F1} & \textbf{Ex.} & \textbf{Str.} & \textbf{Loo.} \\
% \midrule
% Base (+ SFT) & 39.06 & 77.63 & 14.39 & 17.02 \\
% \midrule
% Correctness-only DPO & 39.38 & 73.19 & 15.21 & 17.11 \\
% Chosen-response SFT & 41.75 & 79.52 & 15.04 & 18.09 \\
% Mixed-stage DPO & 42.72 & 81.17 & 15.21 & 17.92 \\
% \midrule
% Only Stage 2 (C-DPO) & 40.27 & 77.88 & 15.13 & 17.43 \\
% Only Stage 1 (M-DPO) & 43.05 & 81.33 & 16.53 & 19.08 \\
% Stage 1 + Stage 2 (Full) & \textbf{45.91} & \textbf{84.79} & \textbf{17.30} & \textbf{20.07} \\
% \bottomrule
% \end{tabular}
% \end{table}

\begin{table}[t]
\centering
\small
\caption{Ablation on training strategy. The first group compares alternative training strategies; the second group ablates individual stages of our pipeline.}
\label{tab:ablation-stage}
\begin{tabular}{l|cccc}
\toprule
\textbf{Strategy} & \textbf{F1} & \textbf{Ex.} & \textbf{Str.} & \textbf{Loo.} \\
\midrule
Base (+ SFT) & 39.06 & 77.63 & 14.39 & 17.02 \\
\midrule
DPO & 39.38 & 73.19 & 15.21 & 17.11 \\
Chosen SFT & 41.75 & 79.52 & 15.04 & 18.09 \\
Mixed DPO & 42.72 & 81.17 & 15.21 & 17.92 \\
\midrule
C-DPO & 40.27 & 77.88 & 15.13 & 17.43 \\
M-DPO & 43.05 & 81.33 & 16.53 & 19.08 \\
Ours(Full) & \textbf{45.91} & \textbf{84.79} & \textbf{17.30} & \textbf{20.07} \\
\bottomrule
\end{tabular}
\end{table}

\subsection{Ablation Studies}
\label{sec:ablation}
All ablation experiments are conducted on Qwen2.5-Math-7B-Instruct.

\paragraph{Effect of Training Strategy.}
We compare against several alternative strategies in Table~\ref{tab:ablation-stage}. 
DPO, the standard preference optimization setup using answer correctness as the sole signal, yields minimal F1 improvement and actually reduces example usage (73.19\% vs.\ 77.63\%), indicating that optimizing solely for correctness can suppress example-based reasoning. 
Chosen SFT, trained on all chosen responses from both stages, achieves moderate gains (F1: 41.75) but falls short of our full pipeline (45.91), confirming that contrastive preference signals outperform imitation alone. 
Mixed DPO, trained on the shuffled union of Stage 1 and Stage 2 preference pairs in a single pass, underperforms even Stage~1 alone (42.72 vs.\ 43.05) despite using more data, suggesting that method and correctness objectives introduce conflicting gradients when mixed. 
Among stage-wise ablations, M-DPO alone (43.05) substantially outperforms C-DPO alone (40.27), confirming that method-oriented training is the critical first step. The full sequential pipeline
consistently yields the best results across all evaluated metrics.

\begin{table}[t]
\centering
\small
\caption{Ablation on response augmentation strategy in 
Stage~1. \textit{Self-sampling} generates candidates 
without guidance; \textit{Teacher rewriting} uses 
DeepSeek-R1 to rewrite references.}
\label{tab:ablation-rgsr}
\begin{tabular}{l|cccc}
\toprule
\textbf{Strategy} & \textbf{F1} & \textbf{Ex.} & \textbf{Str.} & \textbf{Loo.} \\
\midrule
Base (+ SFT)       & 39.06 & 77.63 & 14.39 & 17.02 \\
\midrule
Self-sampling & 40.40 & 78.55 & 15.54 & 17.68 \\
Teacher rewriting  & 39.14 & 74.01 & 15.21 & 17.51 \\
RGSI (Ours)        & \textbf{43.05} & \textbf{81.33} & \textbf{16.53} & \textbf{19.08} \\
\bottomrule
\end{tabular}
\end{table}
\begin{figure}[t]
\centering
\includegraphics[width=\columnwidth]{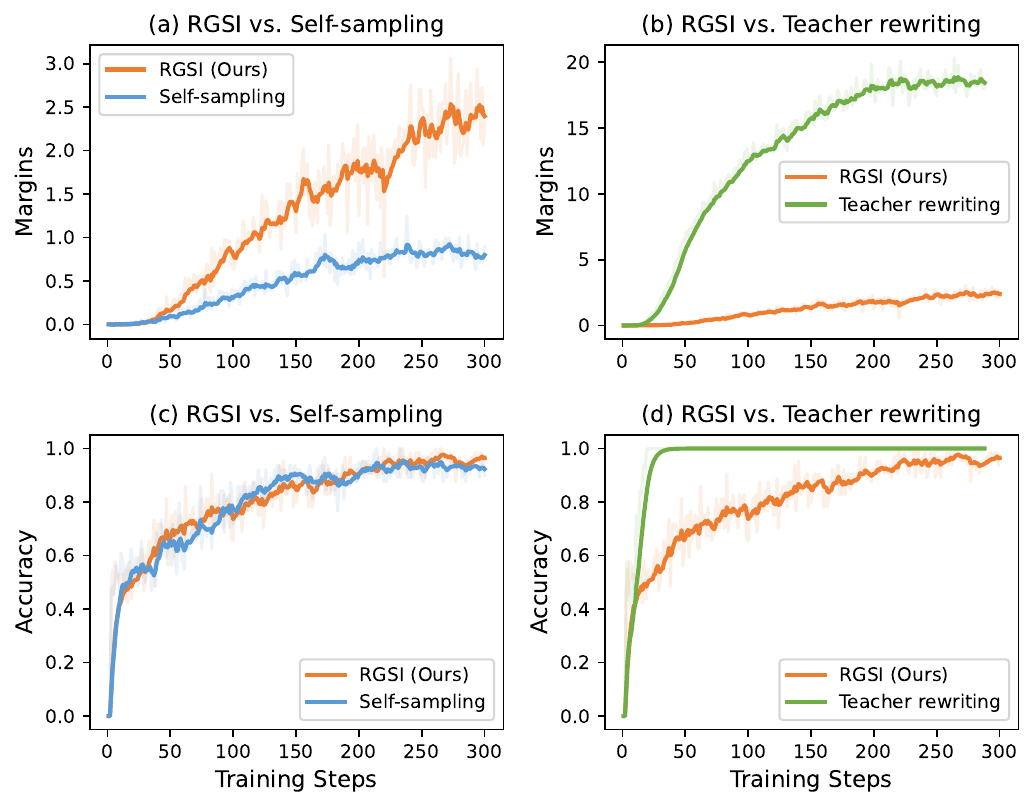}
\caption{DPO training dynamics under different augmentation 
strategies. 
RGSI maintains moderate margins and steady gradual accuracy growth, 
while self-sampling yields insufficient margins and teacher 
rewriting leads to trivially separable pairs.}
\label{fig:rgsr-ablation}
\end{figure}
\paragraph{Effect of Reference-Guided Student Internalization.}
We compare three response augmentation strategies for 
constructing preference pairs in Stage~1: (1) self-sampling,
where all candidates are generated by the base model 
without any guidance; (2) teacher rewriting, where DeepSeek-R1
directly rewrites the reference solutions; 
and (3) our proposed RGSI, where the student model regenerates 
responses after comprehending the reference.

As shown in Table~\ref{tab:ablation-rgsr}, self-sampling 
yields only marginal improvement, as candidates lack sufficient 
quality variation to form effective preference pairs. Teacher 
rewriting, despite producing high-quality responses, nonetheless fails to bring improvement due to distribution shift between the teacher and student generation styles.

Figure~\ref{fig:rgsr-ablation} further illustrates this from 
the training dynamics perspective. Self-sampling results in 
very small margins (a), indicating limited quality differences 
between candidates. Teacher rewriting produces excessively 
large margins (b) and near-instant accuracy saturation (d), 
suggesting the model distinguishes pairs by surface-level 
style rather than reasoning quality. RGSI maintains 
moderate margins (a) with gradual accuracy growth (c), 
providing meaningful learning signals throughout.

\subsection{Analysis}
\label{sec:analysis}
\paragraph{Training vs. Hint Prompting.}
As shown in Table~\ref{tab:hint}, hint prompting increases 
example usage but yields only marginal improvement in 
judgment accuracy (F1: +0.46), suggesting that elicited 
examples are not well-targeted enough to support correct 
reasoning. In contrast, our training approach achieves 
substantially larger gains across all metrics (F1: +7.24), 
indicating the model learns to construct appropriate 
examples that effectively contribute to sound judgments.

\begin{table}[t]
\centering
\small
% \caption{Comparison between hint prompting and our 
% training approach. Hint prompting lets the base model 
% give examples via instruction, while our method learns 
% when and how to use examples through training.}
\caption{Comparison between hint prompting and our stage-wise training approach.}
\label{tab:hint}
\begin{tabular}{l|cccc}
\toprule
\textbf{Setting} & \textbf{F1} & \textbf{Ex.} & \textbf{Str.} & \textbf{Loo.} \\
\midrule
Base                      & 38.67 & 75.25 & 14.31 & 16.61 \\
Base + Hint prompt        & 39.13 & 77.71 & 15.95 & 18.25 \\
Ours (Full)               & \textbf{45.91} & \textbf{84.79} & \textbf{17.30} & \textbf{20.07} \\
\bottomrule
\end{tabular}
\end{table}
\begin{table}[t]
\centering
\small
% \caption{Cross-judge evaluation on Qwen2.5-Math-7B-Instruct. F1 is judge-independent (exact match); Examples, Strict, and Loose Align are evaluated by GPT-4o.}
\caption{Cross-judge evaluation on Qwen2.5-Math-7B-Instruct. F1 is judge-independent; Ex., Str., and Loo. are evaluated by GPT-4o.}
\label{tab:cross-judge-4o}
\begin{tabular}{l|cccc}
\toprule
\textbf{Stage} & \textbf{F1} & \textbf{Ex.} & \textbf{Str.} & \textbf{Loo.} \\
\midrule
Base & 38.67 & 78.21 & 32.81 & 37.83 \\
+ SFT & 39.06 & 79.11 & 33.88 & 38.89 \\
+ Stage 1 & 43.05 & 86.10 & 37.34 & 42.19 \\
+ Stage 1 + Stage 2 & \textbf{45.91} & \textbf{87.50} & \textbf{39.56} & \textbf{43.67} \\
\bottomrule
\end{tabular}
\end{table}
\paragraph{Method Acquisition Analysis.}
To quantify the behavioral shift induced by Stage 1, we compare the method score distribution of sampled responses from $\pi_0$ and $\pi_1$ (Appendix~\ref{app:additional-analysis}). After Stage 1, substantive example usage (m = 2) increases from 30.74\% to 39.30\%, while purely abstract reasoning (m = 0) decreases from 56.58\% to 48.47\%, confirming that method-oriented training shifts the model toward example-based reasoning.

\paragraph{Cross-Judge Validation.}
We adopt DeepSeek-V3.2 as our judge model, favoring an open-source model for reproducibility. To rule out judge bias, we re-evaluate all stages using GPT-4o. As shown in Table~\ref{tab:cross-judge-4o}, although absolute scores shift due to judge calibration, the progressive improvement across stages holds consistently, confirming our results are not an artifact of a specific judge.

\paragraph{Case Study.}
% We provide a qualitative case study in Appendix~\ref{app:additional-analysis}, illustrating the progressive shift from abstract theorem application to targeted counterexample construction across training stages.
A case study in Appendix~\ref{app:additional-analysis} illustrates the progressive shift from abstract reasoning to targeted counterexample construction.

\section{Conclusion}
In this paper, we propose INSPIRE, an Internalize-Then-Improve approach that enhances example-driven mathematical reasoning through Reference-Guided Student Internalization and stage-wise rubric preference training. 
Experiments across multiple model scales and families demonstrate consistent improvements, with maintained or improved performance on out-of-distribution benchmarks. Our results confirm that decomposing capability acquisition into strategy adoption and correctness refinement is more effective than joint optimization. Future work may explore scaling to larger datasets and models, as well as extending the approach to other underexplored reasoning strategies.

%%%%%%%%%%%
\newpage
\section*{Limitations}
Our training data consists of 1,275 problems from BrokenMath, covering a limited set of mathematical domains. Whether the framework generalizes to larger-scale data and broader areas such as number theory and combinatorics remains to be explored. We validate on models up to 7B parameters; the effectiveness on larger scales (e.g., 32B or 72B) has not been examined. Both data construction and evaluation rely on DeepSeek-V3.2 as the judge model, which may introduce biases; we mitigate this through cross-validation with GPT-4o, confirming consistent trends. All reported results are from single runs with fixed random seeds, though the consistent improvements across three model configurations suggest the findings are robust.

\section*{Acknowledgments}

This work was supported in part by the New Generation Artificial
Intelligence--National Science and Technology Major Project
(2025ZD0123003) and the National Natural Science Foundation of China
Enterprise Innovation and Development Joint Fund (Artificial
Intelligence Field) Key Support Projects (U25B2072).

% Our training data consists of 1,275 problems filtered from BrokenMath, covering a limited set of mathematical domains such as algebra, topology, real analysis, and functional analysis. Whether the proposed framework generalizes effectively to larger-scale training data and broader mathematical areas, such as number theory, combinatorics, and probability theory, remains to be explored. 
% Similarly, evaluation is primarily conducted on CounterMath; validating on additional example-oriented benchmarks would further strengthen the generalizability of our findings.

% We validate our approach on 1.5B and 7B scales from the Qwen2.5-Math series and additionally on Llama-3.1-8B-Instruct. While the consistent improvements across these configurations are encouraging, the effectiveness of stage-wise training on larger models (e.g., 32B or 72B) has not been examined, and the training dynamics may differ at those scales. All experiments use fixed random seeds and identical hyperparameters across configurations for reproducibility.

% Both the rubric-based scoring during data construction and the evaluation metrics on CounterMath rely on DeepSeek-V3.2 as the judge model. Although this setup is consistent with the evaluation protocol of the original CounterMath benchmark, inherent biases in the judge model may affect both the quality of constructed preference pairs and the reliability of reported results. We mitigate this concern through cross-validation with GPT-4o, which confirms consistent improvement trends.
% camera-ready时再加

%\section*{Acknowledgments}

\bibliography{custom}

@article{abouelenin2025phi4mini,
  title={Phi-4-Mini Technical Report: Compact yet Powerful Multimodal Language Models via Mixture-of-LoRAs},
  author={Abouelenin, Abdelrahman and Ashfaq, Atabak and Atkinson, Adam and Awadalla, Hany and Bach, Nguyen and Bao, Jianmin and Benhaim, Alon and Cai, Martin and Chaudhary, Vishrav and Chen, Congcong and others},
  journal={Preprint, arXiv:2503.01743},
  year={2025}
}

@article{grattafiori2024llama3,
  title={The Llama 3 Herd of Models},
  author={Grattafiori, Aaron and Dubey, Abhimanyu and Jauhri, Abhinav and Pandey, Abhinav and Kadian, Abhishek and Al-Dahle, Ahmad and Letman, Aiesha and Mathur, Akhil and Schelten, Alan and others},
  journal={Preprint, arXiv:2407.21783},
  year={2024}
}

@misc{anthropic2025claude45sonnet,
  title={Claude Sonnet 4.5 System Card},
  author={{Anthropic}},
  year={2025},
  howpublished={\url{https://www.anthropic.com/claude-sonnet-4-5-system-card}}
}

@misc{google2025gemini3pro,
  title={Gemini 3 Pro Model Card},
  author={{Google DeepMind}},
  year={2025},
  howpublished={\url{https://storage.googleapis.com/deepmind-media/Model-Cards/Gemini-3-Pro-Model-Card.pdf}}
}

@article{zeng2025glm,
  title={GLM-4.5: Agentic, Reasoning, and Coding (ARC) Foundation Models},
  author={Zeng, Aohan and Lv, Xin and Zheng, Qinkai and Hou, Zhenyu and Chen, Bin and Xie, Chengxing and Wang, Cunxiang and Yin, Da and Zeng, Hao and Zhang, Jiajie and others},
  journal={Preprint, arXiv:2508.06471},
  year={2025}
}

@article{bai2026kimik25,
  title={Kimi K2.5: Visual Agentic Intelligence},
  author={Bai, Tongtong and Bai, Yifan and Bao, Yiping and Cai, S. H. and Cao, Yuan and Chen, Cheng and Chen, Guanduo and Chen, Huarong and Chen, Jia and Chen, Jiahao and others},
  journal={Preprint, arXiv:2602.02276},
  year={2026}
}

@article{gunjal2025rubrics,
  title={Rubrics as Rewards: Reinforcement Learning Beyond Verifiable Domains},
  author={Gunjal, Anisha and Wang, Anthony and Lau, Elaine and Nath, Vaskar and He, Yunzhong and Liu, Bing and Hendryx, Sean},
  journal={Preprint, arXiv:2507.17746},
  year={2025}
}

@article{huang2025rubric,
  title={Reinforcement Learning with Rubric Anchors},
  author={Huang, Zenan and Zhuang, Yihong and Lu, Guoshan and Qin, Zeyu and Xu, Haokai and Zhao, Tianyu and Peng, Ru and Hu, Jiaqi and Shen, Zhanming and Hu, Xiaomeng and others},
  journal={Preprint, arXiv:2508.12790},
  year={2025}
}

@inproceedings{wang2024mathshepherd,
  title     = {{Math-Shepherd}: Verify and Reinforce {LLMs} Step-by-step without Human Annotations},
  author    = {Peiyi Wang and Lei Li and Zhihong Shao and Runxin Xu and Damai Dai and Yifei Li and Deli Chen and Yu Wu and Zhifang Sui},
  booktitle = {Proceedings of the 62nd Annual Meeting of the Association for Computational Linguistics (Volume 1: Long Papers)},
  pages     = {9426--9439},
  year      = {2024}
}

@article{singh2024beyond,
  title   = {Beyond Human Data: Scaling Self-Training for Problem-Solving with Language Models},
  author  = {Avi Singh and John D. Co-Reyes and Rishabh Agarwal and Ankesh Anand and Piyush Patil and Xavier Garcia and Peter J. Liu and James Harrison and Jaehoon Lee and Kelvin Xu and others},
  journal = {Transactions on Machine Learning Research},
  year    = {2024}
}

@inproceedings{yuan2024selfrewarding,
  title     = {Self-Rewarding Language Models},
  author    = {Weizhe Yuan and Richard Yuanzhe Pang and Kyunghyun Cho and Sainbayar Sukhbaatar and Jing Xu and Jason Weston},
  booktitle = {Proceedings of the 41st International Conference on Machine Learning},
  year      = {2024}
}

@article{xu2024chatglmmath,
  title   = {{ChatGLM-Math}: Improving Math Problem-Solving in Large Language Models with a Self-Critique Pipeline},
  author  = {Yifan Xu and Xiao Liu and Xinghan Liu and Zhenyu Hou and Yueyan Li and Xiaohan Zhang and Zihan Wang and Aohan Zeng and Zhengxiao Du and Wenyi Zhao and Jie Tang and Yuxiao Dong},
  journal = {Preprint, arXiv:2404.02893},
  year    = {2024}
}

@inproceedings{toshniwal2024openmath,
  title     = {{OpenMathInstruct-1}: A 1.8 Million Math Instruction Tuning Dataset},
  author    = {Shubham Toshniwal and Ivan Moshkov and Sean Narenthiran and Daria Gitman and Fei Jia and Igor Gitman},
  booktitle = {NeurIPS Datasets and Benchmarks},
  year      = {2024}
}

@article{du2025kimik15,
  title   = {Kimi k1.5: Scaling Reinforcement Learning with {LLMs}},
  author  = {Angang Du and Bofei Gao and Bowei Xing and Changjiu Jiang and Cheng Chen and Cheng Li and Chenjun Xiao and Chenzhuang Du and Chonghua Liao and Chuning Tang and others},
  journal = {Preprint, arXiv:2501.12599},
  year    = {2025}
}

@article{ye2025limo,
  title   = {{LIMO}: Less is More for Reasoning},
  author  = {Yixin Ye and Zhen Huang and Yang Xiao and Ethan Chern and Shijie Xia and Pengfei Liu},
  journal = {Preprint, arXiv:2502.03387},
  year    = {2025}
}

@article{ying2024internlmmath,
  title   = {{InternLM-Math}: Open Math Large Language Models Toward Verifiable Reasoning},
  author  = {Huaiyuan Ying and Shuo Zhang and Linyang Li and Zhejian Zhou and Yunfan Shao and Zhaoye Fei and Yichuan Ma and Jiawei Hong and Kuikun Liu and Ziyi Wang and others},
  journal = {Preprint, arXiv:2402.06332},
  year    = {2024}
}

@inproceedings{muennighoff2025s1,
  title     = {s1: Simple Test-Time Scaling},
  author    = {Niklas Muennighoff and Zitong Yang and Weijia Shi and Xiang Lisa Li and Li Fei-Fei and Hannaneh Hajishirzi and Luke Zettlemoyer and Percy Liang and Emmanuel Cand{\`e}s and Tatsunori Hashimoto},
  booktitle = {Proceedings of the 2025 Conference on Empirical Methods in Natural Language Processing},
  year      = {2025}
}

@article{yang2025qwen3,
  title   = {Qwen3 Technical Report},
  author  = {An Yang and Anfeng Li and Baosong Yang and Beichen Zhang and Binyuan Hui and Bo Zheng and Bowen Yu and Chang Gao and Chengen Huang and Chenxu Lv and others},
  journal = {Preprint, arXiv:2505.09388},
  year    = {2025}
}

@article{yang2024qwen25math,
  title   = {Qwen2.5-Math Technical Report: Toward Mathematical Expert Model via Self-Improvement},
  author  = {An Yang and Beichen Zhang and Binyuan Hui and Bofei Gao and Bowen Yu and Chengpeng Li and Dayiheng Liu and Jianhong Tu and Jingren Zhou and Junyang Lin and others},
  journal = {Preprint, arXiv:2409.12122},
  year    = {2024}
}

@article{guo2025deepseekr1,
  title   = {{DeepSeek-R1}: Incentivizing Reasoning Capability in {LLMs} via Reinforcement Learning},
  author  = {Daya Guo and Dejian Yang and Haowei Zhang and Junxiao Song and Peiyi Wang and Runxin Xu and Xiao Bi and Zhan Li and Zhihong Shao and Zhi-Rui Zhang and others},
  journal = {Nature},
  volume  = {645},
  number  = {8081},
  pages   = {633--638},
  year    = {2025}
}

@article{singh2025gpt5,
  title   = {{OpenAI GPT-5} System Card},
  author  = {Aaditya Singh and Adam Fry and Adam Perelman and Adam Tart and Adi Ganesh and Ahmed El-Kishky and Aidan McLaughlin and Aiden Low and AJ Ostrow and Akhila Ananthram and others},
  journal = {Preprint, arXiv:2601.03267},
  year    = {2025}
}

@article{yu2025dapo,
  title   = {{DAPO}: An Open-Source {LLM} Reinforcement Learning System at Scale},
  author  = {Qiying Yu and Zheng Zhang and Ruofei Zhu and Yufeng Yuan and Xiaochen Zuo and Yu Yue and Weinan Dai and Tiantian Fan and Gaohong Liu and Lingjun Liu and others},
  journal = {Preprint, arXiv:2503.14476},
  year    = {2025}
}

@inproceedings{luo2025wizardmath,
  title     = {{WizardMath}: Empowering Mathematical Reasoning for Large Language Models via Reinforced Evol-Instruct},
  author    = {Haipeng Luo and Qingfeng Sun and Can Xu and Pu Zhao and Jian-Guang Lou and Chongyang Tao and Xiubo Geng and Qingwei Lin and Shifeng Chen and Yansong Tang and Dongmei Zhang},
  booktitle = {The Thirteenth International Conference on Learning Representations},
  year      = {2025}
}

@inproceedings{yu2024metamath,
  title     = {{MetaMath}: Bootstrap Your Own Mathematical Questions for Large Language Models},
  author    = {Longhui Yu and Weisen Jiang and Han Shi and Jincheng Yu and Zhengying Liu and Yu Zhang and James Kwok and Zhenguo Li and Adrian Weller and Weiyang Liu},
  booktitle = {The Twelfth International Conference on Learning Representations},
  year      = {2024}
}

@misc{li2024numinamath,
  title        = {{NuminaMath}: The Largest Public Dataset in {AI4Maths} with 860k Pairs of Competition Math Problems and Solutions},
  author       = {Jia Li and Edward Beeching and Lewis Tunstall and Ben Lipkin and Roman Soletskyi and Shengyi Huang and Kashif Rasul and Longhui Yu and Albert Q. Jiang and Ziju Shen and Bin Dong and Li Zhou and Yann Fleureau and Guillaume Lample and Stanislas Polu},
  year         = {2024},
  howpublished = {\url{https://huggingface.co/datasets/AI-MO/NuminaMath-CoT}}
}

@article{shao2024deepseekmath,
  title   = {{DeepSeekMath}: Pushing the Limits of Mathematical Reasoning in Open Language Models},
  author  = {Zhihong Shao and Peiyi Wang and Qihao Zhu and Runxin Xu and Junxiao Song and Xiao Bi and Haowei Zhang and Mingchuan Zhang and Y. K. Li and Y. Wu and Daya Guo},
  journal = {Preprint, arXiv:2402.03300},
  year    = {2024}
}

@article{jaech2024o1,
  title   = {{OpenAI} o1 System Card},
  author  = {Aaron Jaech and Adam Kalai and Adam Lerer and Adam Richardson and Ahmed El-Kishky and Aiden Low and Alec Helyar and Aleksander Madry and Alex Beutel and Alex Carney and others},
  journal = {Preprint, arXiv:2412.16720},
  year    = {2024}
}

@book{klymchuk2010counterexamples,
  title     = {Counterexamples in Calculus},
  author    = {Klymchuk, Sergiy},
  volume    = {34},
  year      = {2010},
  publisher = {American Mathematical Society}
}

@inproceedings{wu2024conceptmath,
  title     = {{ConceptMath}: A Bilingual Concept-Wise Benchmark for Measuring Mathematical Reasoning of Large Language Models},
  author    = {Yanan Wu and Jie Liu and Xingyuan Bu and Jiaheng Liu and Zhanhui Zhou and Yuanxing Zhang and Chenchen Zhang and Zhiqi Bai and Haibin Chen and Tiezheng Ge},
  booktitle = {Findings of the Association for Computational Linguistics: ACL 2024},
  pages     = {6815--6839},
  year      = {2024}
}

@inproceedings{li2025one,
  title     = {One Example Shown, Many Concepts Known! Counterexample-Driven Conceptual Reasoning in Mathematical {LLM}s},
  author    = {Yinghui Li and Jiayi Kuang and Haojing Huang and Zhikun Xu and Xinnian Liang and Yi Yu and Wenlian Lu and Yangning Li and Xiaoyu Tan and Chao Qu and Ying Shen and Haitao Zheng and Philip S. Yu},
  booktitle = {Forty-second International Conference on Machine Learning},
  year      = {2025}
}

@inproceedings{rafailov2023dpo,
  title     = {Direct Preference Optimization: Your Language Model is Secretly a Reward Model},
  author    = {Rafael Rafailov and Archit Sharma and Eric Mitchell and Christopher D. Manning and Stefano Ermon and Chelsea Finn},
  booktitle = {Advances in Neural Information Processing Systems},
  volume    = {36},
  pages     = {53728--53741},
  year      = {2023}
}

@inproceedings{azar2024general,
  title     = {A General Theoretical Paradigm to Understand Learning from Human Preferences},
  author    = {Mohammad Gheshlaghi Azar and Zhaohan Daniel Guo and Bilal Piot and Remi Munos and Mark Rowland and Michal Valko and Daniele Calandriello},
  booktitle = {International Conference on Artificial Intelligence and Statistics},
  pages     = {4447--4455},
  year      = {2024}
}

@inproceedings{meng2024simpo,
  title     = {{SimPO}: Simple Preference Optimization with a Reference-Free Reward},
  author    = {Yu Meng and Mengzhou Xia and Danqi Chen},
  booktitle = {Advances in Neural Information Processing Systems},
  volume    = {37},
  pages     = {124198--124235},
  year      = {2024}
}

@article{lai2024stepdpo,
  title   = {{Step-DPO}: Step-wise Preference Optimization for Long-Chain Reasoning of {LLMs}},
  author  = {Xin Lai and Zhuotao Tian and Yukang Chen and Senqiao Yang and Xiangru Peng and Jiaya Jia},
  journal = {Preprint, arXiv:2406.18629},
  year    = {2024}
}

@inproceedings{xu2025fullstepdpo,
  title     = {{Full-Step-DPO}: Self-Supervised Preference Optimization with Step-Wise Rewards for Mathematical Reasoning},
  author    = {Huimin Xu and Xinnian Mao and Feng-Lin Li and Xiaobao Wu and Wang Chen and Wei Zhang and Luu Anh Tuan},
  booktitle = {Findings of the Association for Computational Linguistics: ACL 2025},
  pages     = {24343--24356},
  year      = {2025}
}

@inproceedings{zheng2023judging,
  title     = {Judging {LLM}-as-a-Judge with {MT-Bench} and Chatbot Arena},
  author    = {Lianmin Zheng and Wei-Lin Chiang and Ying Sheng and Siyuan Zhuang and Zhanghao Wu and Yonghao Zhuang and Zi Lin and Zhuohan Li and Dacheng Li and Eric Xing and others},
  booktitle = {Advances in Neural Information Processing Systems},
  volume    = {36},
  pages     = {46595--46623},
  year      = {2023}
}

@article{petrov2025brokenmath,
  title   = {{BrokenMath}: A Benchmark for Sycophancy in Theorem Proving with {LLMs}},
  author  = {Ivo Petrov and Jasper Dekoninck and Martin Vechev},
  journal = {Preprint, arXiv:2510.04721},
  year    = {2025}
}

@article{cui2025prime,
  title   = {Process Reinforcement through Implicit Rewards},
  author  = {Ganqu Cui and Lifan Yuan and Zefan Wang and Hanbin Wang and Yuchen Zhang and Jiacheng Chen and Wendi Li and Bingxiang He and Yuchen Fan and Tianyu Yu and others},
  journal = {Preprint, arXiv:2502.01456},
  year    = {2025}
}

@inproceedings{zheng2024llamafactory,
  title     = {{LlamaFactory}: Unified Efficient Fine-Tuning of 100+ Language Models},
  author    = {Yaowei Zheng and Richong Zhang and Junhao Zhang and Yanhan Ye and Zheyan Luo and Zhangchi Feng and Yongqiang Ma},
  booktitle = {Proceedings of the 62nd Annual Meeting of the Association for Computational Linguistics},
  year      = {2024}
}

@inproceedings{kwon2023vllm,
  title     = {Efficient Memory Management for Large Language Model Serving with {PagedAttention}},
  author    = {Woosuk Kwon and Zhuohan Li and Siyuan Zhuang and Ying Sheng and Lianmin Zheng and Cody Hao Yu and Joseph E. Gonzalez and Hao Zhang and Ion Stoica},
  booktitle = {Proceedings of the 29th Symposium on Operating Systems Principles},
  year      = {2023}
}

@inproceedings{hu2022lora,
  title     = {{LoRA}: Low-Rank Adaptation of Large Language Models},
  author    = {Edward J. Hu and Yelong Shen and Phillip Wallis and Zeyuan Allen-Zhu and Yuanzhi Li and Shean Wang and Lu Wang and Weizhu Chen},
  booktitle = {International Conference on Learning Representations},
  year      = {2022}
}

@article{liu2024deepseekv3,
  title   = {{DeepSeek-V3} Technical Report},
  author  = {Aixin Liu and Bei Feng and Bing Xue and Bingxuan Wang and Bochao Wu and Chengda Lu and Chenggang Zhao and Chengqi Deng and Chenyu Zhang and Chong Ruan and others},
  journal = {Preprint, arXiv:2412.19437},
  year    = {2024}
}

@article{li2026learning,
  title={Learning to Disprove: Formal Counterexample Generation with Large Language Models},
  author={Zenan Li and Zhaoyu Li and Kaiyu Yang and Xiaoxing Ma and Zhendong Su},
  journal={Preprint, arXiv:2603.19514},
  year={2026}
}

@inproceedings{kuang2025atomic,
  title={Atomic Thinking of LLMs: Decoupling and Exploring Mathematical Reasoning Abilities},
  author={Jiayi Kuang and Haojing Huang and Yinghui Li and Xinnian Liang and Zhikun Xu and Yangning Li and Xiaoyu Tan and Chao Qu and Meishan Zhang and Ying Shen and Philip S. Yu},
  booktitle={Advances in Neural Information Processing Systems},
  year={2025}
}

@inproceedings{li2022past,
  title={The past mistake is the future wisdom: Error-driven contrastive probability optimization for chinese spell checking},
  author={Li, Yinghui and Zhou, Qingyu and Li, Yangning and Li, Zhongli and Liu, Ruiyang and Sun, Rongyi and Wang, Zizhen and Li, Chao and Cao, Yunbo and Zheng, Hai-Tao},
  booktitle={Findings of the Association for Computational Linguistics: ACL 2022},
  pages={3202--3213},
  year={2022}
}

@article{li2024llms,
  title={When llms meet cunning texts: A fallacy understanding benchmark for large language models},
  author={Li, Yinghui and Zhou, Qingyu and Luo, Yuanzhen and Ma, Shirong and Li, Yangning and Zheng, Hai-Tao and Hu, Xuming and Yu, Philip S},
  journal={Advances in Neural Information Processing Systems},
  volume={37},
  pages={112433--112458},
  year={2024}
}

@inproceedings{li2025rethinking,
  title={Rethinking the roles of large language models in chinese grammatical error correction},
  author={Li, Yinghui and Qin, Shang and Ye, Jingheng and Huang, Haojing and Li, Yangning and Guo, Shu-Yu and Qin, Libo and Hu, Xuming and Jiang, Wenhao and Zheng, Hai-Tao and others},
  booktitle={Proceedings of the 63rd Annual Meeting of the Association for Computational Linguistics (Volume 6: Industry Track)},
  pages={553--567},
  year={2025}
}

@article{li2025refine,
  title={Refine knowledge of large language models via adaptive contrastive learning},
  author={Li, Yinghui and Huang, Haojing and Kuang, Jiayi and Li, Yangning and Guo, Shu-Yu and Qu, Chao and Tan, Xiaoyu and Zheng, Hai-Tao and Shen, Ying and Yu, Philip S},
  journal={arXiv preprint arXiv:2502.07184},
  year={2025}
}

@article{lu2025youtu,
  title={Youtu-llm: Unlocking the native agentic potential for lightweight large language models},
  author={Lu, Junru and Qin, Jiarui and Qiao, Lingfeng and Li, Yinghui and Dai, Xinyi and Ke, Bo and He, Jianfeng and Qiao, Ruizhi and Yin, Di and Sun, Xing and others},
  journal={arXiv preprint arXiv:2512.24618},
  year={2025}
}

@article{xu2026topoagent,
  title={TopoAgent: A Self-Evolving Topological Agent for Multimodal Scientific Reasoning},
  author={Xu, Mingze and Li, Yinghui and Kuang, Jiayi and Kang, Zhanhui and Yin, Di and Shen, Ying and Sun, Xing and Han, Yuxing},
  journal={arXiv preprint arXiv:2607.14658},
  year={2026}
}

@article{zhang2026chatbot,
  title={From Chatbot to Digital Colleague: The Paradigm Shift Toward Persistent Autonomous AI},
  author={Zhang, Yongheng and Liu, Ziang and Zhu, Jiaxuan and Wang, Shuai and Chen, Xiangqi and Huang, Haojing and Kuang, Jiayi and Chen, Siyu and Shen, Ao and Wu, Hao and others},
  journal={arXiv preprint arXiv:2606.14502},
  year={2026}
}

@article{an2026toward,
  title={Toward Native Multimodal Modeling: A Roadmap},
  author={An, Siyu and Lu, Junru and Dong, Junnan and Wang, Qiufeng and Li, Yinghui and Fei, Weizhi and Yu, Zichao and Yuan, Zheng and Liu, Biao and Wang, Haopeng and others},
  journal={arXiv preprint arXiv:2605.25343},
  year={2026}
}

@article{li2026cognitive,
  title={Cognitive Mismatch in Multimodal Large Language Models for Discrete Symbol Understanding},
  author={Li, Yinghui and Kuang, Jiayi and Xing, Peng and Liu, Daixian and Zhang, Yongheng and Dong, Junnan and Guo, Shu-Yu and Li, Yangning and Zhou, Qingyu and Jiang, Wenhao and others},
  journal={arXiv preprint arXiv:2603.18472},
  year={2026}
}

@article{shen2026deep,
  title={Deep Thought Alignment: Trajectory-Level Latent Distillation for Video Reasoning},
  author={Shen, Ao and Zhang, Yongheng and Li, Yinghui and Wang, Manning and Yin, Di and Sun, Xing},
  journal={arXiv preprint arXiv:2608.16316},
  year={2026}
}

@inproceedings{kuang2026process,
  title={Process-level trajectory evaluation for environment configuration in software engineering agents},
  author={Kuang, Jiayi and Li, Yinghui and Zhang, Xin and Li, Yangning and Sun, Xing and Shen, Ying and Yu, Philip and others},
  booktitle={International Conference on Learning Representations},
  volume={2026},
  pages={113832--113855},
  year={2026}
}

@article{liu2026tangrampuzzle,
  title={TangramPuzzle: Evaluating Multimodal Large Language Models with Compositional Spatial Reasoning},
  author={Liu, Daixian and Kuang, Jiayi and Li, Yinghui and Li, Yangning and Yin, Di and Cao, Haoyu and Sun, Xing and Shen, Ying and Zheng, Hai-Tao and Lin, Liang and others},
  journal={arXiv preprint arXiv:2601.16520},
  year={2026}
}
% \bibliographystyle{acl_natbib}

% 正文结束后
\newpage
\onecolumn
\appendix

\section{Evaluation Rubric Prompts}
\label{app:rubric}
We provide the prompt used for scoring candidate 
responses along the Method and Reasoning dimensions 
defined in Table~\ref{tab:rubric}. Each response is 
scored independently by the judge model 
using the following prompt. The Correctness dimension 
is determined by exact matching between the model's 
final answer and the ground-truth label.

\begin{figure}[H]
\centering
\includegraphics[width=\textwidth]{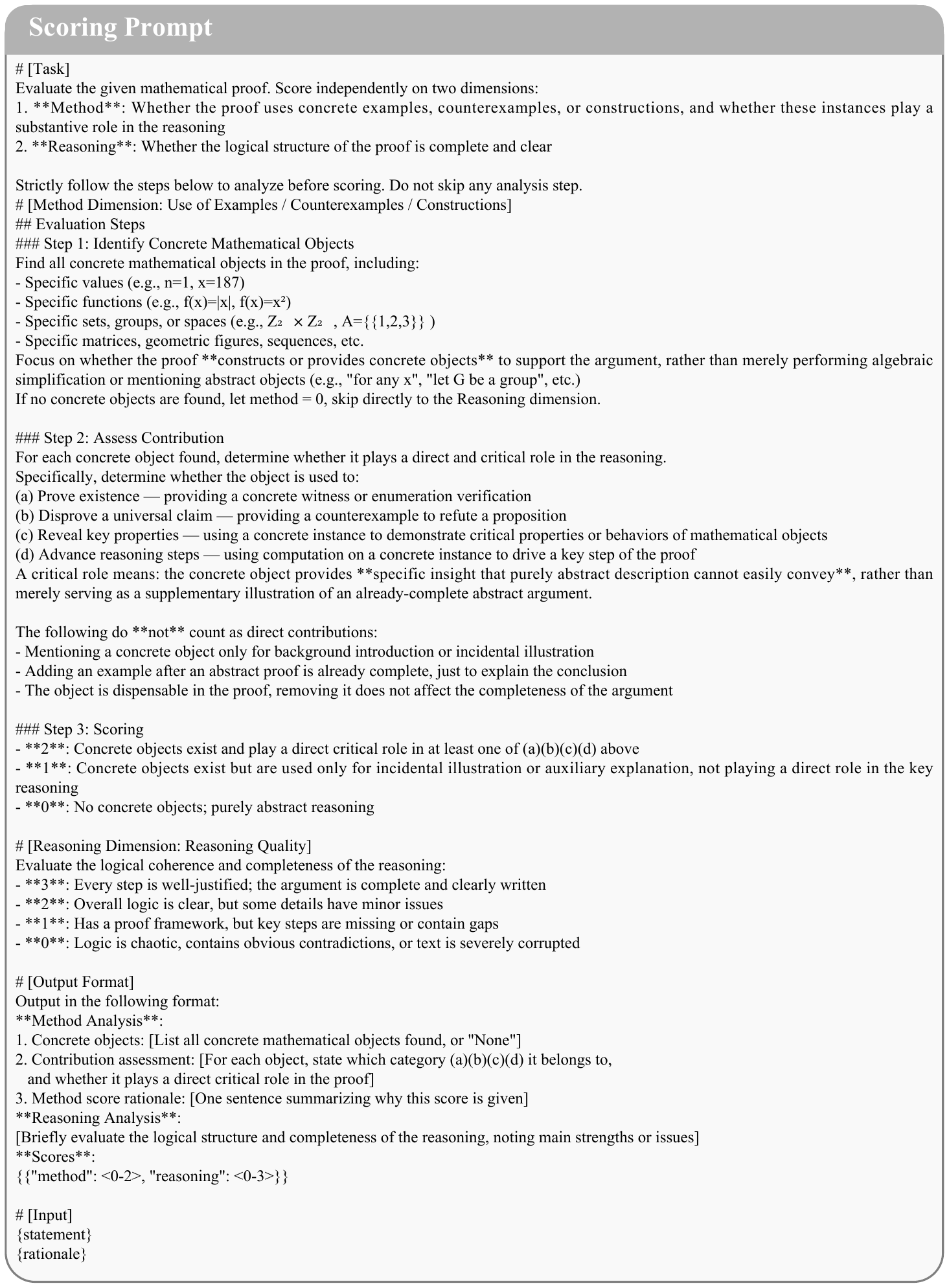}
\end{figure}

\section{Problem Filtering Criteria}
\label{app:filtering}

We use DeepSeek-R1 to identify candidate problems 
from BrokenMath that involve examples, counterexamples, 
or constructions. The filtering prompt is shown below.

\begin{figure}[H]
\centering
\includegraphics[width=\textwidth]{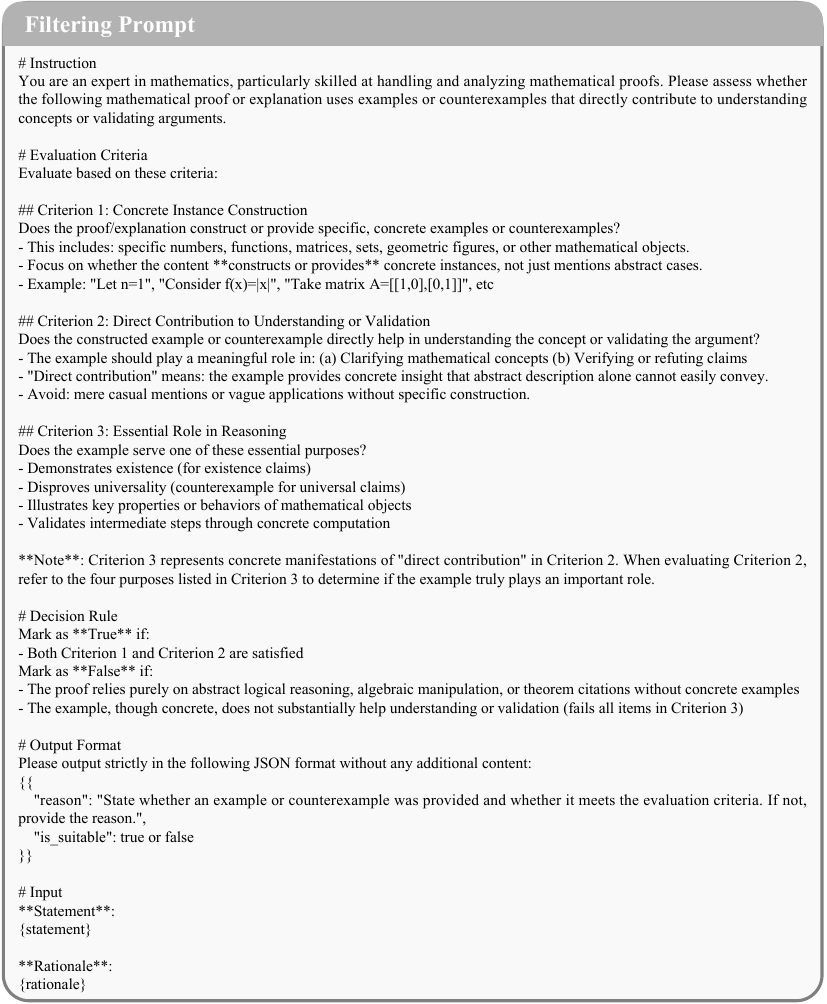}
\end{figure}

\newpage
\section{Reference-Guided Student Internalization}
\label{app:rgsi}

In RGSI, the base model is provided with the reference 
solution and asked to regenerate a complete response in 
its own expression style. The rewriting prompt is shown below.

\begin{figure}[H]
\centering
\includegraphics[width=\textwidth]{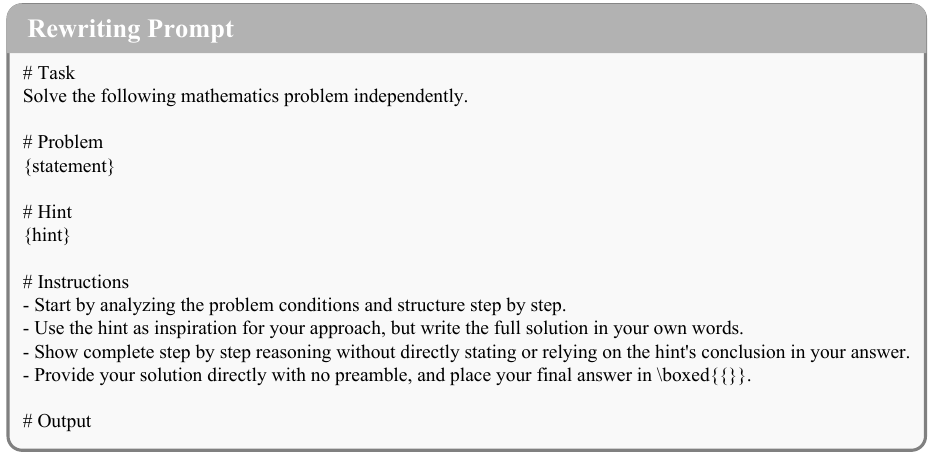}
\end{figure}

\section{Inference Prompt Templates}
\label{app:inference}
We use two inference prompts in our evaluation. 
The standard prompt is used for all main experiments, 
and the hint prompt explicitly requests example-based 
reasoning, as compared in Section~\ref{sec:analysis}.

\begin{figure}[H]
\centering
\includegraphics[width=\textwidth]{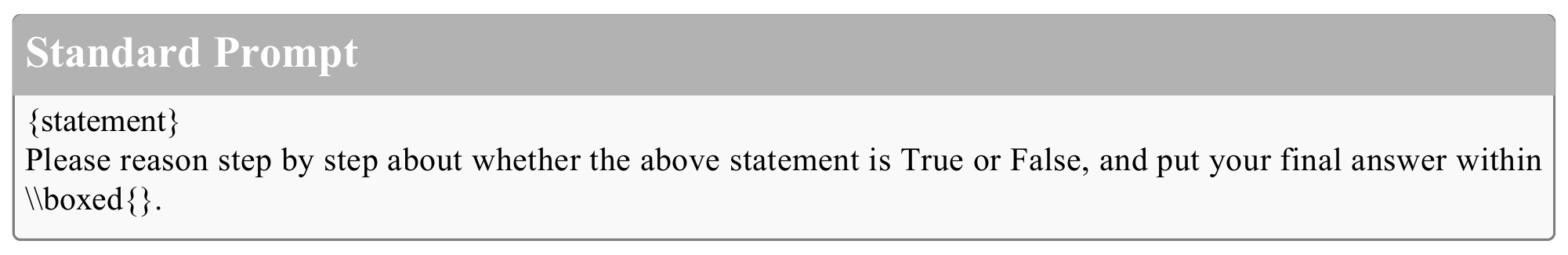}
\end{figure}

\begin{figure}[H]
\centering
\includegraphics[width=\textwidth]{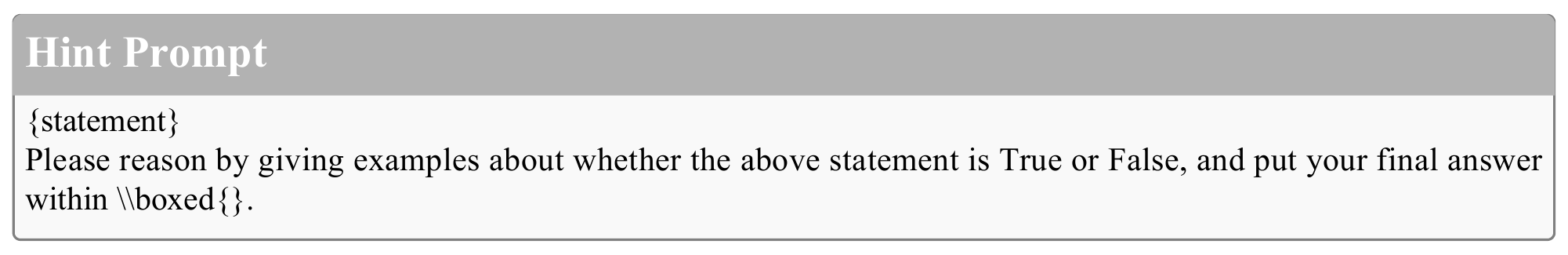}
\end{figure}

\section{Training and Details}
\label{app:eval-details}
\paragraph{Evaluation Metrics.}
We adopt the evaluation metrics from CounterMath~\citep{li2025one}: 
\textbf{Macro-F1} measures the accuracy of the model's true/false 
judgments on mathematical statements, using macro averaging to 
account for label imbalance; 
\textbf{Examples} measures the proportion of problems where the 
model employs examples during reasoning; 
\textbf{Strict Align} measures the proportion of model-provided 
examples that are fully consistent with the reference; 
\textbf{Loose Align} measures the proportion of instances where 
at least one model-provided example is consistent with the reference. 
We use DeepSeek-V3.2 as the judge model for computing these metrics, 
with the same evaluation prompts as CounterMath.

\paragraph{Hyperparameters.}
For SFT, we set the learning rate to $1 \times 10^{-5}$ and train 
for 3 epochs. For DPO, we set the learning rate to 
$5 \times 10^{-6}$, $\beta = 0.2$, and train for 3 epochs. 
The same hyperparameters are used for both model scales.

\section{Additional Analysis}
\label{app:additional-analysis}

\paragraph{Method Score Distribution.}
For each of the 1,275 problems, we sample 16 responses from both
$\pi_0$ and $\pi_1$ under the same settings, yielding 20,400
self-sampled responses per model without RGSI responses. After
DeepSeek-V3.2 scoring and parsing, 20,245 responses from $\pi_0$ and
20,171 from $\pi_1$ remain valid (155 and 229 excluded, respectively).
Table~\ref{tab:method-dist} reports the resulting distributions.

\begin{table}[H]
\centering
\small
\caption{Method-score distribution of self-sampled responses before and after Stage 1.}
\label{tab:method-dist}
\resizebox{\columnwidth}{!}{%
\begin{tabular}{lrrrr}
\toprule
\textbf{Model}
& \textbf{Valid responses}
& $m=0$
& $m=1$
& $m=2$ \\
\midrule
$\pi_0$ (before Stage 1)
& 20,245
& 11,455 (56.58\%)
& 2,567 (12.68\%)
& 6,223 (30.74\%) \\
$\pi_1$ (after Stage 1)
& 20,171
& 9,777 (48.47\%)
& 2,466 (12.23\%)
& 7,928 (39.30\%) \\
\bottomrule
\end{tabular}%
}
\end{table}

After Stage 1, substantive example usage ($m=2$) increases from
30.74\% to 39.30\%, while purely abstract reasoning ($m=0$) decreases
from 56.58\% to 48.47\%.

\begin{figure*}[h]
\centering
\includegraphics[width=1\textwidth]{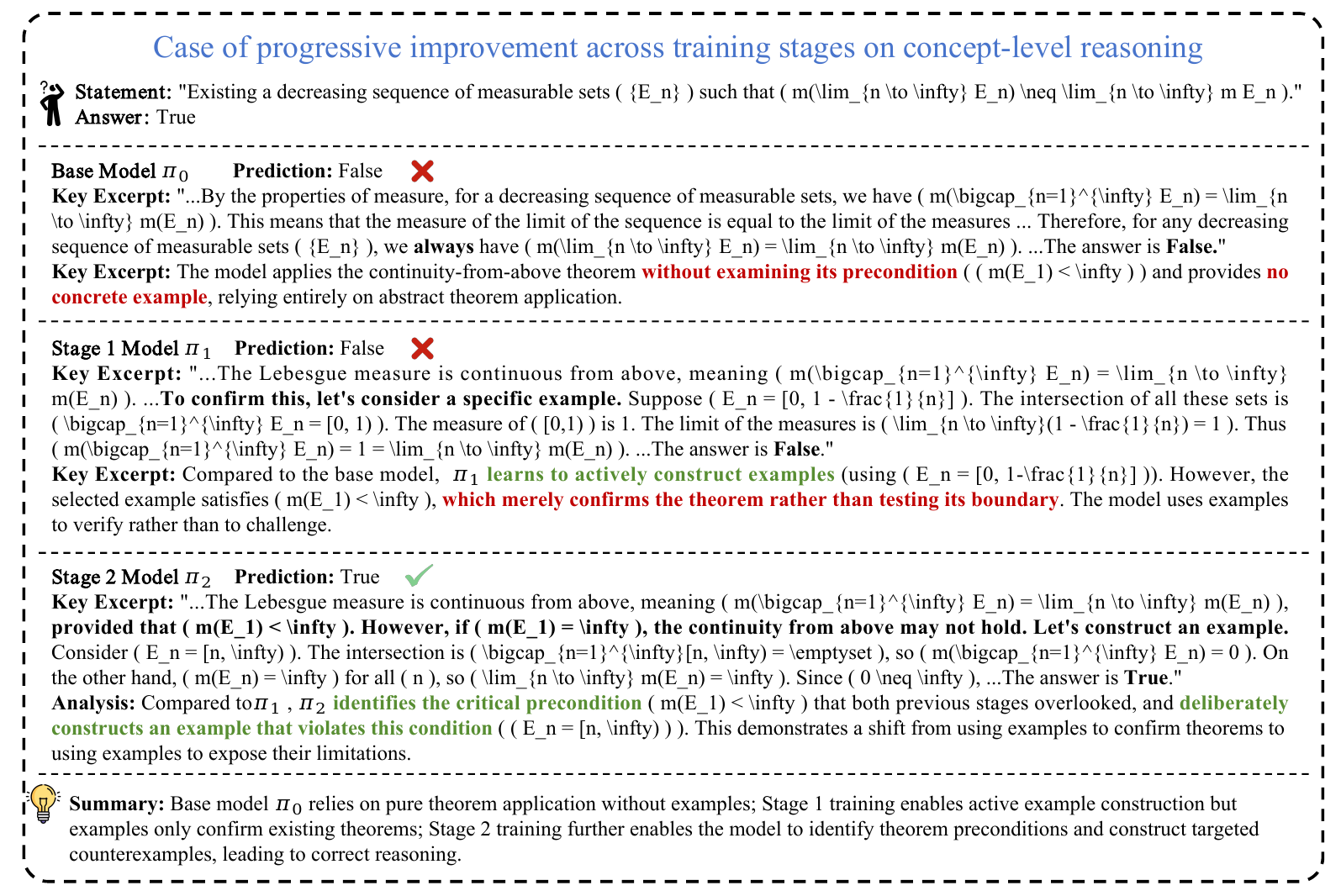}
\caption{Case study comparing reasoning across training stages on a measure theory problem from CounterMath.}
\label{fig:case-study}
\end{figure*}

\paragraph{Case Study.}
Figure~\ref{fig:case-study} presents a representative example 
from CounterMath. The base model $\pi_0$ applies the 
continuity-from-above theorem without examining its 
precondition ($m(E_1) < \infty$), leading to an incorrect 
judgment through purely abstract reasoning. After Stage~1 
training, $\pi_1$ learns to actively construct examples, 
but the chosen example satisfies the theorem's precondition 
and merely confirms rather than challenges it. After Stage~2 
training, $\pi_2$ identifies the critical precondition and 
deliberately constructs a counterexample that violates it 
($E_n = [n, \infty)$), arriving at the correct answer. This 
progression demonstrates that Stage~1 equips the model with 
the \emph{strategy} of using examples, while Stage~2 refines 
the \emph{quality} by teaching it to construct targeted 
examples that test theorem boundaries.

%%%%% 
\newpage
\section{Detailed Rubric Criteria}
\label{app:rubric_detail}

Table~\ref{tab:rubric_detail} provides the full scoring criteria for each dimension and score level defined in our evaluation rubric (Table~\ref{tab:rubric}).

\begin{table*}[h]
\centering
\small
\begin{tabular}{llp{10.5cm}}
\toprule
\textbf{Dimension} & \textbf{Score} & \textbf{Criteria} \\
\midrule
\multirow{3}{*}{Method ($m$)}
 & 2 & Concrete mathematical objects (specific values, functions, sets, matrices, etc.) are constructed and play a \textbf{direct critical role} in the reasoning: proving existence, disproving a universal claim, revealing key properties, or advancing a key reasoning step, et al. \\
 & 1 & Concrete objects are present but serve only as incidental illustration or auxiliary explanation, without playing a direct role in the core argument. \\
 & 0 & No concrete objects; the response relies entirely on abstract reasoning or theorem application. \\
\midrule
\multirow{4}{*}{Reasoning ($r$)}
 & 3 & Every step is well-justified; the argument is complete and clearly written. \\
 & 2 & Overall logic is clear, but some details have minor issues. \\
 & 1 & Has a proof framework, but key steps are missing or contain logical gaps. \\
 & 0 & Logic is chaotic, contains obvious contradictions, or text is severely corrupted. \\
\midrule
\multirow{2}{*}{Correctness ($c$)}
 & 1 & The final answer matches the ground-truth label. \\
 & 0 & The final answer does not match the ground-truth label. \\
\bottomrule
\end{tabular}
\caption{Detailed scoring criteria for each dimension and score level of the evaluation rubric.}
\label{tab:rubric_detail}
\end{table*}

\section{Additional Experiments and Reproducibility Details}
\label{app:additional_details}

\subsection{Evaluation on a Recent Backbone}
\label{app:recent_backbone}

To examine whether INSPIRE remains effective on a more recent
backbone, we apply the full two-stage pipeline to
Qwen3-4B-Instruct-2507\cite{yang2025qwen3} following the same training and evaluation
pipeline as in our main experiments.
\begin{table}[H]
\centering
\small
\caption{Evaluation of INSPIRE on the recent Qwen3-4B-Instruct-2507 backbone. Bold indicates the best result.}
\label{tab:qwen3-recent-backbone}
\begin{tabular}{l|cccc}
\toprule
\textbf{Stage} & \textbf{F1} & \textbf{Ex.} & \textbf{Str.} & \textbf{Loo.} \\
\midrule
Base
& 33.65 & 65.21 & 18.83 & 25.65 \\
+ SFT
& 33.71 & 65.79 & 19.32 & 26.39 \\
+ Stage 1
& 35.35 & 73.03 & 20.23 & 27.06 \\
+ Stage 1 + Stage 2
& \textbf{36.36} & \textbf{74.34} & \textbf{21.96} & \textbf{29.03} \\
\bottomrule
\end{tabular}
\end{table}

\subsection{Correctness-Only DPO Sensitivity}
\label{app:dpo_sensitivity}

We train the correctness-only DPO baseline from the same SFT
checkpoint and candidate pool as our main pipeline, obtaining 10,539
preference pairs based solely on final-answer correctness. All variants
use LoRA with a learning rate of $5\times10^{-6}$ for three epochs,
while $\beta$ is varied with all other settings fixed.

\begin{table}[H]
\centering
\small
\caption{Sensitivity of the correctness-only DPO baseline to $\beta$ on Qwen2.5-Math-7B-Instruct. Bold indicates the best result.}
\label{tab:dpo-beta-sensitivity}
\begin{tabular}{l|cccc}
\toprule
\textbf{Setting} & \textbf{F1} & \textbf{Ex.} & \textbf{Str.} & \textbf{Loo.} \\
\midrule
Base (+ SFT)
& 39.06 & 77.63 & 14.39 & 17.02 \\

DPO ($\beta=0.1$)
& 38.90 & 73.27 & 15.87 & 17.51 \\

DPO ($\beta=0.2$)
& 39.38 & 73.19 & 15.21 & 17.11 \\

DPO ($\beta=0.5$)
& 38.94 & 74.42 & 13.56 & 15.71 \\
\midrule
Ours (Full)
& \textbf{45.91} & \textbf{84.79} & \textbf{17.30} & \textbf{20.07} \\
\bottomrule
\end{tabular}
\end{table}

Correctness-only DPO remains close to the SFT baseline across all
values of $\beta$ and consistently reduces example usage, suggesting
that its limited effectiveness is not specific to a particular
regularization setting.

\end{document}